\documentclass[11pt]{article}
\usepackage[letterpaper,margin=1in]{geometry}
\usepackage{times}
\usepackage{natbib}
\setcitestyle{authoryear,round,citesep={;},aysep={,},yysep={;}}
\usepackage{graphicx}
\usepackage{float}

\usepackage{amsmath,amsfonts,bm}

\def\eqref#1{equation~\ref{#1}}

\def\1{\bm{1}}

\DeclareMathAlphabet{\mathsfit}{\encodingdefault}{\sfdefault}{m}{sl}
\SetMathAlphabet{\mathsfit}{bold}{\encodingdefault}{\sfdefault}{bx}{n}

\usepackage{multirow}
\usepackage{hyperref}
\hypersetup{hidelinks,
  pdfauthor={Di Zhang, Zhangpeng Gong, Jiashuai Liu, Zhi Zeng, Jiusong Ge, Chunze Yang, Xitong Ling, Kai Yi, Kai He, Weimiao Yu, Mireia Crispin-Ortuzar, Chen Li, Zeyu Gao},
  pdftitle={Can Protein-Derived Knowledge Improve Pathology Foundation Models?}}
\usepackage{url}
\usepackage{booktabs}
\usepackage{pifont}
\usepackage{longtable}
\usepackage{tabularx}
\title{Can Protein-Derived Knowledge Improve Pathology Foundation Models?}
\date{}

\author{\begin{minipage}{\dimexpr\textwidth-12pt\relax}\centering\small
Di Zhang\textsuperscript{1}, Zhangpeng Gong\textsuperscript{1},
Jiashuai Liu\textsuperscript{1}, Zhi Zeng\textsuperscript{1}\\
Jiusong Ge\textsuperscript{1}, Chunze Yang\textsuperscript{1},
Xitong Ling\textsuperscript{2}, Kai Yi\textsuperscript{3}, Kai He\textsuperscript{4}\\
Weimiao Yu\textsuperscript{5}, Mireia Crispin-Ortuzar\textsuperscript{6},
Chen Li\textsuperscript{7,1,*}, Zeyu Gao\textsuperscript{6,*}\\[0.7em]
\textsuperscript{1}School of Computer Science and Technology, Xi'an Jiaotong University\\
\textsuperscript{2}Tsinghua Shenzhen International Graduate School, Tsinghua University\\
\textsuperscript{3}University of Cambridge\\
\textsuperscript{4}Saw Swee Hock School of Public Health, National University of Singapore\\
\textsuperscript{5}Bioinformatics Institute (BII), A*STAR\\
\textsuperscript{6}Department of Oncology, University of Cambridge\\
\textsuperscript{7}Institute of Intelligent Medicine, Peking Union Medical College\\[0.4em]
\textsuperscript{*}Co-corresponding authors:\\
Chen Li (\href{mailto:cli@pumc.edu.cn}{cli@pumc.edu.cn});
Zeyu Gao (\href{mailto:zg323@cam.ac.uk}{zg323@cam.ac.uk})
\end{minipage}}
\begin{document}

\maketitle
\begin{abstract}
Molecularly guided pathology foundation models (PFMs) exploit transcriptomic or proteomic information to enrich whole-slide image (WSI) representations, yet effectively leveraging large standalone molecular corpora remains challenging. First, existing molecular foundation models encode protein sequences or single-cell states, not the patient-level bulk expression profiles paired with WSIs. Second, because cross-modal supervision is restricted to paired WSI–omics samples, knowledge from standalone molecular corpora reaches the pathology encoder only indirectly, creating a paired-support bottleneck. To address these challenges, we propose a three-stage framework that decouples proteomic knowledge acquisition from cross-modal transfer, yielding ProSlide, a slide-level hierarchical pathology foundation model. First, to close the modality gap, we pretrain a Proteomic Foundation Encoder (PFE) on 12,695 sample-level bulk protein profiles using virtual profile generation and expression-space multi-view pretraining. Second, we pretrain ProSlide, a patch–region–slide encoder, to predict protein expression from paired WSI–protein samples. Third, to relax the paired-support bottleneck, we introduce Prot2Path, a cross-modal relational distillation objective. For each paired sample, it aligns the similarity distributions of the WSI and its protein profile over a shared, frozen bank of PFE-encoded paired and standalone profiles. We evaluate ProSlide on 12 downstream tasks across breast, lung, and renal cancers. Despite being pretrained with only 2,229 WSIs and 12,695 sample-level protein profiles, ProSlide achieves the highest mean accuracy and AUC within each cancer group.

\end{abstract}


\section{Introduction}
Foundation models have advanced computational pathology by learning transferable representations from large-scale histopathology data \citep{gao2026alpaca,uni}, supporting diverse diagnostic \citep{gao2025smmile,li2025mico}, prognostic \citep{yang2026path}, and molecular prediction tasks \citep{lffa}. More recently, pathology foundation models (PFMs) have extended morphology-driven pretraining by incorporating complementary molecular supervision \citep{xu2025multimodal}. Molecular profiles provide an additional view of disease biology, offering supervision that connects tissue morphology with underlying molecular characteristics. Multimodal frameworks such as Threads \citep{threads} and CARE \citep{zhang2026care} exemplify this direction by integrating histopathological and molecular information to enrich whole-slide image (WSI) representations. These advances highlight the value of biological information for pathology representation learning. 

\begin{figure}[t]
\begin{center}
\includegraphics[width=\linewidth]{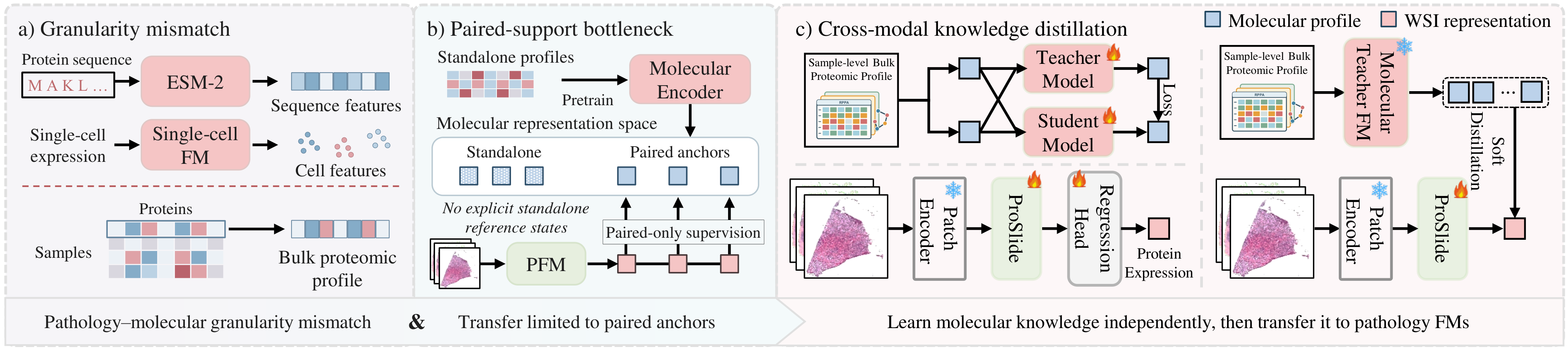}
\end{center}
\caption{(a) Granularity mismatch between existing molecular encoders and sample-level proteomic states. (b) Cross-modal transfer is restricted to paired WSI–omics anchors. (c) Our framework leverages standalone molecular profiles through cross-modal knowledge distillation.}
\label{motivation}
\end{figure}

However, finding the optimal strategy to exploit such molecular knowledge for pathology pretraining is challenging. Existing molecularly guided PFMs largely rely on paired WSI–omics cohorts for cross-modal learning \citep{threads,zhang2026care}. As shown in Fig.~\ref{motivation}(a), the biological units behind typical molecular encoders are much smaller in physical scale than a WSI. For example, protein language models \citep{esm} primarily encode protein identity and sequence-derived properties, and single-cell models \citep{cui2023scGPT} characterize cellular expression states. In contrast, the proteomic measurements associated with WSIs normally represent sample-level bulk protein abundance patterns. 
This motivates learning a sample-level proteomic representation space aligned with the molecular states used for pathology supervision.

A second challenge arises when transferring such molecular knowledge to pathology. Even when a molecular encoder is pretrained on a substantially larger collection of standalone profiles, conventional cross-modal learning typically queries this representation space only at molecular samples with corresponding WSIs \citep{zhang2026care}. As illustrated in Fig.~\ref{motivation}(b), standalone profiles can shape the molecular encoder indirectly during pretraining, but they do not explicitly participate in the pathology training objective. Consequently, the pathology encoder is supervised only at a limited set of paired molecular anchors and is not directly constrained to preserve its relationships to molecular states outside the paired cohort. This creates a paired-support bottleneck between large-scale molecular representation learning and cross-modal pathology training.

To address these limitations, we introduce ProSlide, a hierarchical pathology encoder trained through a three-stage framework that separates proteomic knowledge acquisition from cross-modal transfer (Fig.~\ref{motivation}(c)). First, we pretrain a Proteomic Foundation Encoder (PFE) on sample-level bulk protein profiles using expression-space multi-view pretraining and virtual profile generation. This stage establishes a molecular representation space by learning from proteomic cohorts independently of WSI pairing. Second, we pretrain ProSlide to predict protein expression from paired WSI–protein samples, using region-level attention to aggregate patch features into slide representations. Third, we introduce Prot2Path to transfer PFE-defined inter-sample relations to ProSlide, complementing the direct supervision provided by protein-expression prediction. Prot2Path constructs a frozen bank of PFE embeddings from both paired and standalone protein profiles as a shared molecular reference. 
Standalone profiles thus contribute through two routes: shaping the teacher’s representation space during PFE pretraining and providing molecular references during cross-modal distillation. 
We evaluate PFE against a BERT-style pretrained counterpart through profile retrieval under partial protein masking and downstream pathology distillation. Our default PFE configuration achieves higher retrieval accuracy and better downstream performance, supporting its utility as a
proteomic teacher for pathology representation learning. We further evaluate ProSlide on 12 tasks, with four each for breast, lung, and renal cancers. Despite being pretrained with only 2,229 WSIs and 12,695 sample-level protein profiles, ProSlide achieves the highest mean accuracy and AUC within each cancer group.
Our main contributions are summarized as follows:

\noindent\textbf{(1) Independent proteomic representation learning.}
We introduce \textbf{PFE}, which learns sample-level proteomic representations through expression-space multi-view pretraining and virtual profile generation without requiring paired WSIs.

\noindent\textbf{(2) Cross-modal relational distillation.}
We develop \textbf{ProSlide}, a protein-supervised hierarchical pathology encoder, and train it with \textbf{Prot2Path} to match PFE-defined similarity distributions over a shared bank of paired and standalone protein profiles.

\noindent\textbf{(3) Effective learning under limited pairing.}
Across 12 downstream tasks, ProSlide achieves the highest mean performance within each of three cancer groups using limited paired training data.
\section{Related Work}
\paragraph{Molecularly Guided Pathology Representation Learning.}
Pathology foundation models learn transferable representations from large histology collections \citep{uni,li20252}, while hierarchical encoders such as HIPT aggregate patch features into slide representations \citep{hipt}. Molecularly guided approaches extend this paradigm using paired WSI--omics data. Threads incorporates genomic and transcriptomic supervision \citep{threads}, and CARE uses RNA and protein profiles to refine adaptive region representations \citep{zhang2026care}. Our framework complements these approaches by independently pretraining a proteomic encoder on standalone data before transferring its knowledge to ProSlide through paired samples.

\paragraph{Protein and Proteomic Representation Learning.}
Protein language models such as ESM-2 learn sequence representations that support structure prediction \citep{esm}. At the expression level, scPROTEIN uses graph contrastive learning to generate single-cell proteomic embeddings \citep{li2024scprotein}. PFE targets bulk protein expression profiles, where each sample contains continuous measurements of multiple identified proteins. Building on multi-view self-supervised learning \citep{dinov2}, PFE combines virtual profile generation with identity-preserving expression perturbations to learn proteomic representations for subsequent pathology distillation.

\paragraph{Knowledge Distillation.}
Knowledge distillation transfers teacher predictions or representations to a student, while relational variants transfer relationships among samples \citep{park2019relational}. In pathology, Pathryoshka consolidates multiple pretrained pathology teachers into a compact model \citep{grashei2025pathryoshka}. Cross-modal approaches such as MoMKD transfer genomic supervision through a momentum-updated memory for histology-only inference \citep{guo2026momentum}. Prot2Path uses an independently pretrained proteomic teacher and a frozen bank containing paired and standalone profiles. It transfers proteomic relations by aligning protein- and WSI-derived similarity distributions over the same sampled references.

\section{Method}
\begin{figure}[t]
\begin{center}
\includegraphics[width=\linewidth]{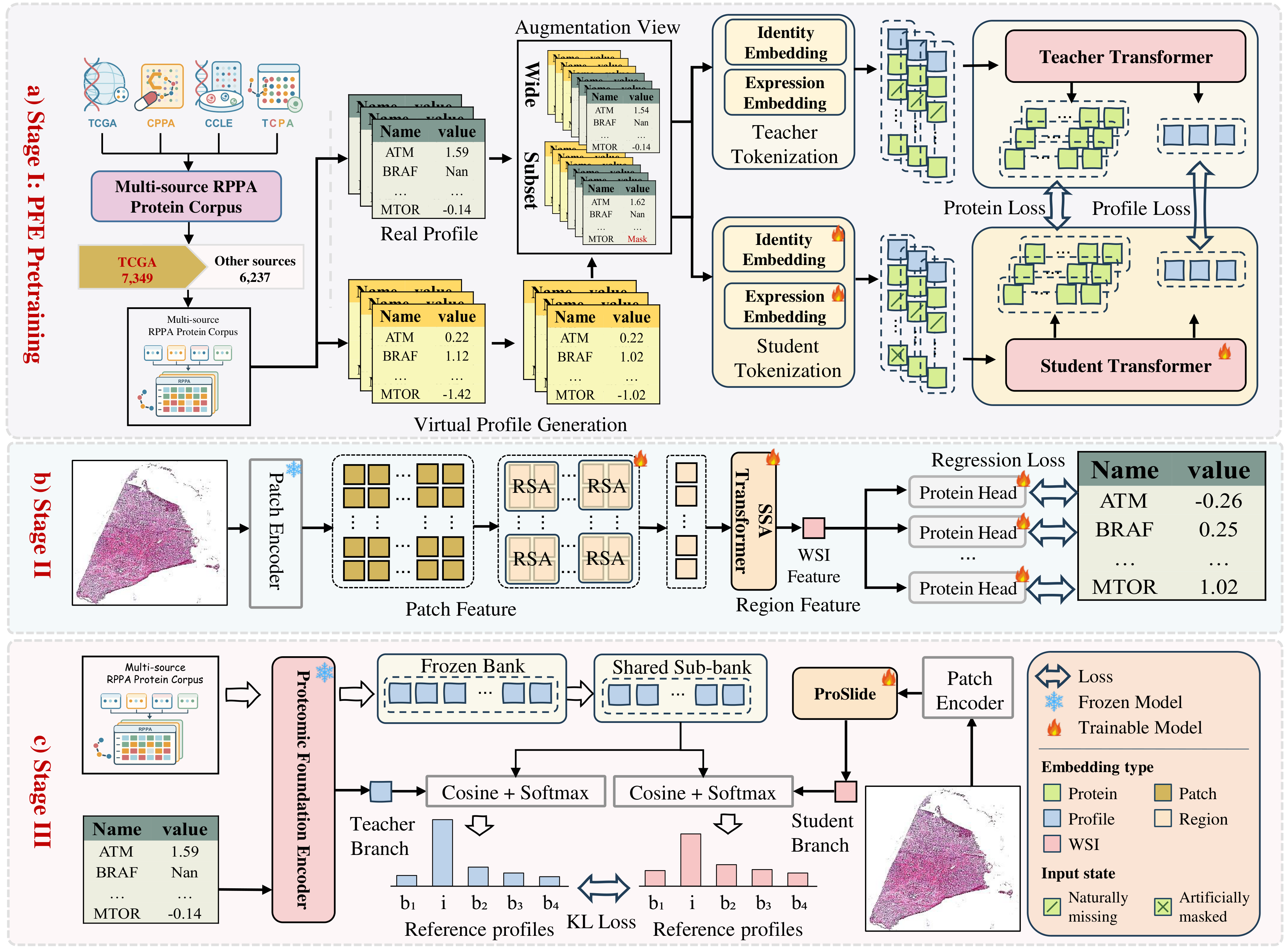}
\end{center}
\caption{\textbf{Overview of the three-stage framework.} (a) PFE learns proteomic representations through virtual profile generation and expression-space multi-view pretraining. (b) ProSlide learns hierarchical WSI representations through protein-expression prediction. (c) Prot2Path distills proteomic relational knowledge into ProSlide through similarity distribution alignment over a shared sampled protein bank.}
\label{framework}
\end{figure}

Our framework uses standalone proteomic data for both proteomic representation learning and cross-modal knowledge transfer through three training stages (Fig.~\ref{framework}). First, we pretrain a Proteomic Foundation Encoder (PFE) on sample-level protein profiles independently of WSI pairing. Second, we pretrain ProSlide to predict protein expression from paired WSI–protein samples. Third, Prot2Path transfers inter-sample relations from the frozen PFE to ProSlide through a shared representation bank containing both paired and standalone protein profiles.
\subsection{Stage I: Proteomic Foundation Encoder Pretraining}
\label{sec:pfe}
To support subsequent cross-modal transfer, we establish a proteomic representation space that captures variation across sample-level bulk protein profiles. PFE combines protein identity with continuous expression values to represent each profile. We train the encoder using expression-space multi-view pretraining, with complementary objectives operating at the profile and protein levels.

\subsubsection{PFE Architecture}

We curate RPPA-based bulk protein expression profiles from four resources, TCGA \citep{akbani2014pan}, TCPA \citep{li2013tcpa}, CCLE \citep{ghandi2019next}, and
CPPA \citep{zhao2020large}, covering sample tumor specimens, baseline cancer cell lines, and pharmacologically perturbed cell-line samples (Appendix~\ref{app:protein_data}). Each profile is represented as
\begin{equation}
\mathcal{X}_i = \{(p_j, x_{i,j}, m_{i,j})\}_{j=1}^{M},
\label{eq:profile}
\end{equation}
where $i$ indexes samples and $j$ indexes proteins. Here, $p_j$ denotes protein identity, $x_{i,j}\in\mathbb{R}$ is the normalized expression value, and $m_{i,j}\in\{0,1\}$ indicates whether the measurement is observed.
PFE constructs each protein token using a learnable identity embedding $e_j^{\mathrm{id}}\in\mathbb{R}^{d}$ and a shared value encoder $g_{\mathrm{val}}:\mathbb{R}\rightarrow\mathbb{R}^{d}$:
\begin{equation}
e_{i,j} = e_j^{\mathrm{id}} + m_{i,j}g_{\mathrm{val}}(x_{i,j}) + (1-m_{i,j})e^{\mathrm{mask}},
\label{protein_token}
\end{equation}
where $e^{\mathrm{mask}}\in\mathbb{R}^{d}$ is a shared learnable embedding for missing measurements.

An $L$-layer Transformer processes these tokens with a prepended learnable \texttt{[CLS]} token, without positional embeddings. Its outputs $h_{i,0}$ and $\{h_{i,j}\}_{j=1}^{M}$ provide the profile-level and contextualized protein-level representations, respectively.

\subsubsection{Expression-Space Multi-View Pretraining}
Inspired by multi-view self-supervised learning \citep{dinov2}, we construct proteomic views by augmenting expression values while preserving protein identities. Our strategy combines virtual profile generation to diversify training inputs with asymmetric teacher--student learning over profile-wide and subset-perturbed views.

\paragraph{Virtual profile generation.}
Each mini-batch contains $B_{\mathrm{r}}$ real profiles from $\mathcal{D}=\{\mathcal{X}_i\}_{i=1}^{N}$ and $B_{\mathrm{v}}$ virtual profiles generated online. For each virtual profile $a$, we independently sample a source profile $\mathcal{X}_{k_a}$ from the full corpus. We select a small subset $\mathcal{S}_a \subseteq\{j:m_{k_a,j}=1\}$ of observed proteins as candidates for expression perturbation:
\begin{equation}
\widetilde{\mathcal{X}}_a = \{(p_j,\tilde{x}_{a,j},m_{k_a,j})\}_{j=1}^{M},
\qquad
\tilde{x}_{a,j} =
\begin{cases}
\mathcal{T}_{a,j}(x_{k_a,j}), & j\in\mathcal{S}_a,\\
x_{k_a,j}, & \text{otherwise}.
\end{cases}
\label{eq:virtual_profile}
\end{equation}
Here, $\mathcal{T}_{a,j}$ denotes the action on protein $j$ of the augmentation sampled for virtual profile $a$. We sample one of three augmentation types: \emph{partial posterization} reduces expression precision, \emph{partial Gaussian perturbation} adds independent noise, and \emph{partial mixup} interpolates values with another profile. Mixup modifies only proteins observed in both profiles, using one mixing coefficient shared across the interpolated coordinates. Each virtual-profile augmentation modifies only a small subset of observed proteins, preserving the source profile’s protein identities, missingness pattern, and unselected expression values while introducing additional local expression configurations for pretraining. Operator definitions are provided in Appendix~\ref{app:pretrain_details}.


\paragraph{Expression-space multi-view construction.}
For each real or virtual profile, we construct profile-wide and subset-perturbed views. Both retain all $M$ protein identities and preserve naturally missing measurements.

\textit{Profile-wide views.} We construct three views for each profile: one clean view and two stochastically augmented views. The augmented views apply mild posterization to all observed expression values, introducing variations in numerical precision while retaining the complete protein context.

\textit{Subset-perturbed views.} Additional views apply Gaussian noise and random masking to independently sampled subsets of observed proteins. Random masking hides expression values while retaining protein identity tokens, and unselected expression values remain unchanged.

The resulting views are used to train PFE with complementary profile- and protein-level self-supervised objectives following DINOv2. View assignment, objective definitions, and training configurations are provided in Appendix~\ref{app:protein_pretrain}.

\subsection{Stage II: Protein-Supervised ProSlide Pretraining}
\label{stage2}
ProSlide adopts a patch-to-region-to-slide hierarchy, operating on pre-extracted features from the frozen CONCH v1.5 patch encoder \citep{conch}. Its region and slide Transformers are randomly initialized and jointly trained to predict protein expression from paired WSI--protein data. This stage provides a molecularly informed initialization for subsequent Prot2Path distillation.

\paragraph{Hierarchical slide encoder.}
For slide $i$, the input is represented as $\mathcal{W}_i = \{(h_{i,n},u_{i,n})\}_{n=1}^{N_i}$, where $N_i$ is the number of patches, $h_{i,n}\in\mathbb{R}^{768}$ denotes the feature of patch $n$, and $u_{i,n}\in\mathbb{R}^{2}$ denotes the patch-grid coordinate of patch $n$ in the WSI. We group patches into $R_i$ non-overlapping spatial regions, with $\mathcal{I}_{i,k}$ denoting the patch indices in region $k$. Within each region, a region self-attention (RSA) module aggregates patch features into a region representation:
\begin{equation}
r_{i,k} = \operatorname{RSA}\left(c^r, \{h_{i,n}\}_{n\in\mathcal{I}_{i,k}}\right),
\label{eq:proslide_region}
\end{equation}
where $c^r$ is a learnable region \texttt{[CLS]} token. RSA projects the patch features into the region embedding space, prepends $c^r$, and processes the resulting sequence with a region Transformer. The output corresponding to $c^r$ serves as the region representation $r_{i,k}$.
The region representations are further aggregated by a slide self-attention (SSA) module to obtain the WSI representation:
\begin{equation}
z_i^{\mathrm{slide}} =
\operatorname{SSA}\left(
c^s,
\{r_{i,k}\}_{k=1}^{R_i}
\right),
\label{eq:proslide_slide}
\end{equation}
where $c^s$ is a learnable slide \texttt{[CLS]} token. SSA projects the region representations into the slide embedding space, prepends $c^s$, and models interactions across regions using a slide Transformer. The output corresponding to $c^s$ serves as the slide representation $z_i^{\mathrm{slide}}\in\mathbb{R}^{768}$.
Both RSA and SSA are implemented as Transformer encoders with multi-head self-attention. RSA models interactions among patches within each region, whereas SSA models interactions among regions across the slide. Region partitioning and architectural configurations are detailed in the Appendix \ref{ProSlideArchitecture}.

\paragraph{Protein-expression supervision.}
A protein-specific regression head $g_j$ predicts the expression of protein $j$ as $\widehat{y}_{i,j}=g_j(z_i^{\mathrm{slide}})$. We minimize masked mean-squared error:
\begin{equation}
\mathcal{L}_{\mathrm{Pro}} =
\frac{\sum_{i\in\mathcal{B}}\sum_j m_{i,j}\left(\widehat{y}_{i,j}-y_{i,j}\right)^2}
{\sum_{i\in\mathcal{B}}\sum_j m_{i,j}},
\label{eq:proslide_loss}
\end{equation}
where $\mathcal{B}$ is a mini-batch of paired samples, $j$ indexes the supervised protein panel, $y_{i,j}$ is the measured expression value, and $m_{i,j}$ indicates whether that measurement is available. This objective trains ProSlide to capture morphological features associated with individual protein expression levels. After pretraining, the regression heads are removed, and the ProSlide encoder provides the initialization for Prot2Path distillation in Stage III.

\subsection{Stage III: Prot2Path Cross-Modal Knowledge Distillation}
\label{stage3}
To further incorporate the proteomic structure learned by PFE, we introduce Prot2Path, which aligns the similarity distributions of paired protein and WSI representations over a shared proteomic representation bank. This relational objective transfers PFE-defined neighborhoods to ProSlide and allows standalone protein profiles to serve as molecular references.

\paragraph{Shared proteomic representation bank.}
We freeze PFE and use it to encode a reference set $\mathcal{D}_{\mathrm{bank}}=\{\mathcal{X}_b\}_{b=1}^{K}$ containing both paired and standalone protein profiles. For each reference profile, we obtain its normalized PFE representation as
\begin{equation}
t_b = \operatorname{PFE}(\mathcal{X}_b),
\qquad
\widehat{t}_b = \frac{t_b}{\lVert t_b\rVert_2}.
\end{equation}
The resulting representations form a fixed proteomic representation bank
\begin{equation}
T =
\left[
\widehat{t}_1;
\widehat{t}_2;
\ldots;
\widehat{t}_K
\right]
\in\mathbb{R}^{K\times d},
\label{eq:protein_bank}
\end{equation}
where $d$ is the PFE embedding dimension. The bank is computed once and remains frozen throughout distillation.

For each paired sample $(\mathcal{W}_i,\mathcal{X}_i)$, we construct a reference index set $\mathcal{J}_i$ containing all paired training profiles and randomly sampled standalone profiles, with $|\mathcal{J}_i|=K_{\mathrm{sub}}$. The sampled sub-bank is defined as
$T_{\mathcal{J}_i}=\left[\widehat{t}_b\right]_{b\in\mathcal{J}_i}\in\mathbb{R}^{K_{\mathrm{sub}}\times d}.$
The index set $\mathcal{J}_i$ is resampled across epochs, allowing the same paired sample to be compared with different proteomic references during training.
PFE defines the proteomic relational target over the sampled sub-bank:
\begin{equation}
\pi_i^{\mathrm{prot}}
=
\operatorname{Softmax}
\left(
\frac{\widehat{t}_i T_{\mathcal{J}_i}^{\top}}
{\tau_{\mathrm{t}}}
\right).
\label{eq:protein_relation}
\end{equation}
where $\tau_{\mathrm{t}}$ is the teacher temperature. Each element of $\pi_i^{\mathrm{prot}}$ measures the relative proteomic similarity between the paired profile $\mathcal{X}_i$ and one reference profile in the bank. Consequently, profiles without corresponding WSIs can still participate in defining the relational target.

\paragraph{Cross-modal relational distillation.}
Given the paired WSI $\mathcal{W}_i$, ProSlide produces the slide representation $z_i^{\mathrm{slide}}$. An alignment head $g_{\phi}$ maps this representation into the PFE embedding space:
\begin{equation}
s_i = g_{\phi}\left(z_i^{\mathrm{slide}}\right),
\qquad
\widehat{s}_i = \frac{s_i}{\lVert s_i\rVert_2}.
\label{eq:pathology_projection}
\end{equation}
Using the same sampled sub-bank, we construct the pathology-derived relational distribution as
\begin{equation}
\pi_i^{\mathrm{path}}
=
\operatorname{Softmax}
\left(
\frac{\widehat{s}_i T_{\mathcal{J}_i}^{\top}}
{\tau_{\mathrm{s}}}
\right).
\label{eq:pathology_relation}
\end{equation}
where $\tau_{\mathrm{s}}$ is the student temperature. 
Because $\pi_i^{\mathrm{prot}}$ and $\pi_i^{\mathrm{path}}$ are defined over the same reference profiles, their corresponding elements describe relations to identical bank entries.

\paragraph{Knowledge distillation loss.}
Prot2Path minimizes the divergence between the proteomic and pathology-derived distributions:
\begin{equation}
\mathcal{L}_{\mathrm{Prot2Path}} =
\frac{1}{|\mathcal{B}|}
\sum_{i\in\mathcal{B}}
D_{\mathrm{KL}}
\left(
\pi_i^{\mathrm{prot}}
\,\Vert\,
\pi_i^{\mathrm{path}}
\right),
\label{eq:prot2path_loss}
\end{equation}
where $\mathcal{B}$ denotes a mini-batch of paired samples. This objective trains ProSlide to reproduce the PFE-defined proteomic neighborhood of each paired sample. The graded target probabilities preserve similarities to both paired and standalone reference profiles.

During distillation, PFE and the representation bank remain frozen, while the alignment head and ProSlide are optimized. After training, PFE, the bank, and the alignment head are discarded. The distilled ProSlide encoder is retained for WSI-only downstream inference. Additional details of the Prot2Path reference bank and distillation objective are provided in the Appendix \ref{Prot2PathImplementation}.

\begin{table*}[t]
\centering
\caption{Logistic-regression results averaged within the selected renal cell carcinoma (RCC), breast, and lung task groups. Each entry reports the mean of the four task-level means $\pm$ the arithmetic mean of their reported standard deviations.}
\label{tab:pancancer-12task-group-average-lr}
\setlength{\tabcolsep}{5pt}
\begin{tabular}{@{}l*{6}{c}@{}}
\toprule
Model
& \multicolumn{2}{c}{Renal}
& \multicolumn{2}{c}{Breast}
& \multicolumn{2}{c}{Lung} \\
\cmidrule(lr){2-3}
\cmidrule(lr){4-5}
\cmidrule(lr){6-7}
& ACC & AUC & ACC & AUC & ACC & AUC \\
\midrule
mean-pooling
& 57.82$\pm$0.25
& 72.16$\pm$0.80
& 67.74$\pm$0.87
& 75.70$\pm$1.01
& 59.61$\pm$0.20
& 75.31$\pm$0.20 \\

FEATHER
& 54.95$\pm$0.44
& 70.96$\pm$1.01
& 66.79$\pm$0.92
& 74.80$\pm$0.82
& 60.22$\pm$0.22
& 74.78$\pm$0.26 \\

GigaPath
& 56.04$\pm$0.25
& 75.33$\pm$0.72
& 67.19$\pm$0.84
& 77.50$\pm$1.08
& 58.93$\pm$0.35
& 73.43$\pm$0.27 \\

PRISM
& 56.28$\pm$0.36
& 72.72$\pm$0.73
& 66.39$\pm$0.88
& 74.88$\pm$0.90
& 61.83$\pm$0.16
& 74.86$\pm$0.17 \\

CARE
& 58.43$\pm$0.31
& 75.24$\pm$0.69
& 67.63$\pm$0.67
& 75.07$\pm$0.83
& 61.94$\pm$0.20
& \underline{76.99$\pm$0.16} \\

TITAN
& \underline{58.67$\pm$0.45}
& \underline{75.74$\pm$0.54}
& \underline{68.84$\pm$0.66}
& \underline{78.19$\pm$0.72}
& \underline{62.18$\pm$0.26}
& 76.15$\pm$0.19 \\

ProSlide
& \textbf{60.85$\pm$0.46}
& \textbf{76.67$\pm$0.61}
& \textbf{70.02$\pm$0.78}
& \textbf{78.53$\pm$0.87}
& \textbf{63.17$\pm$0.21}
& \textbf{77.90$\pm$0.24} \\
\bottomrule
\end{tabular}
\end{table*}

\begin{table*}[t]
\centering
\caption{
Comparison of protein models on internal and external masked-protein retrieval tasks.
The internal results are evaluated on 891 held-out TCGA samples, while the external results are evaluated using the combined 275-patient CPTAC gallery. All retrieval results are reported as the mean R@1 (\%) over five runs.
}
\label{tab:protein_model_mask_retrieval}
\resizebox{\textwidth}{!}{
\begin{tabular}{lccc ccc cc}
\toprule
\multirow{2}{*}{Model}
& \multicolumn{3}{c}{Mask Results}
& \multicolumn{3}{c}{External Mask Results}
& \multicolumn{2}{c}{Distillation Results} \\
\cmidrule(lr){2-4}
\cmidrule(lr){5-7}
\cmidrule(lr){8-9}
& M1 R@1
& M2 R@1
& M3 R@1
& M1 R@1
& M2 R@1
& M3 R@1
& ACC
& AUC \\
\midrule
Protein-BERT & 97.62 & 94.25 & 91.40
   & 96.22 & 92.15 & 84.87
   & 62.64 & 75.99 \\

PFE \& $B_{\mathrm{v}} = 0$ & 97.42 & 93.89 & 89.90
   & 95.35 & 91.71 & 83.35
   & 63.21 & 76.94 \\

PFE \& $B_{\mathrm{v}} = 32$ & 98.47 & 96.36 & 93.76
   & \textbf{97.09} & 94.55 & \textbf{88.51}
   & 64.57 & 77.13 \\

\textbf{PFE \& $B_{\mathrm{v}} = 64$} & \textbf{98.92} & \textbf{96.92} & \textbf{95.17}
   & 96.58 & \textbf{95.05} & 88.44
   & \textbf{64.68} & 77.70 \\
PFE \& $B_{\mathrm{v}} = 128$ & 96.14 & 87.77 & 75.29
   & 94.25 & 88.00 & 75.35
   & 64.57 & \textbf{78.13} \\

\bottomrule
\end{tabular}
}
\end{table*}

\section{Experiments}
\subsection{Experimental Setup}
\label{expsetting}
\paragraph{Datasets and downstream tasks.}
We evaluate our models on 12 WSI-level classification tasks drawn from eight
cancer cohorts. These include four CPTAC cohorts \citep{edwards2015cptac}, CPTAC-BRCA, CPTAC-LUAD, CPTAC-LSCC, and CPTAC-CCRCC, together with the BCNB \citep{xu2021predicting},
Local-LUNG, DHMC-LUNG \citep{wei2019pathologist}, and MUT-HETRCC \citep{acosta2022intratumoral} cohorts. The benchmark contains four tasks each for breast, lung, and renal cancers, covering molecular prediction,
histological subtyping, and cross-cohort transfer. WSI pretraining in Stages II and III uses only TCGA data, whereas all downstream evaluations are conducted on non-TCGA cohorts. Detailed task definitions, cohort statistics, and split protocols are provided in
Appendix~\ref{app:Datasets}.

\paragraph{Implementation details.}
We evaluate frozen representation quality through linear probing, training a logistic regression classifier on fixed slide representations. For the within-cohort CPTAC tasks, we follow the train/test-only evaluation protocol of \citet{threads} and use 50 repeated patient-level 80\%/20\% train/test splits without a separate validation set. For these tasks, logistic regression uses $L_2$ regularization ($C=0.5$), an intercept term, and the L-BFGS solver, with a maximum of 10,000 optimization
iterations. The remaining tasks use five patient-level repetitions with separate training, validation, and test partitions. Task-specific cohort compositions and splitting protocols are provided in Appendix~\ref{app:Datasets}. Test partitions are reserved for final evaluation. All downstream experiments are conducted on a system equipped with a single NVIDIA RTX 4090 GPU.

\paragraph{Evaluation metrics.}
We evaluate all downstream tasks using balanced accuracy and the area under the receiver-operating-characteristic curve (AUROC). Balanced accuracy measures the mean recall across classes and is therefore robust to class imbalance. For multi-class tasks, we report macro-averaged one-versus-rest AUROC. All results are reported as mean $\pm$ standard deviation.

\subsection{Main Results}
To evaluate the quality and transferability of the learned slide representations, we compare ProSlide with mean pooling and five representative pathology foundation models: FEATHER \citep{feather}, GigaPath \citep{gigapath}, PRISM \citep{prism}, CARE \citep{zhang2026care}, and TITAN \citep{titan}. All models are evaluated using frozen slide representations under the same logistic-regression protocol. Table~\ref{tab:pancancer-12task-group-average-lr} summarizes the results averaged over four tasks each for renal, breast, and lung cancers, with complete task-level results provided in Appendix Table~\ref{tab:pancancer-12task-lr}.
ProSlide achieves the highest mean ACC and AUC across all three cancer groups. It performs consistently well across renal mutation-prediction tasks, while the relative strengths of different models vary across breast and lung molecular endpoints.
ProSlide also performs well on pooled and cross-cohort fine-grained lung subtyping, supporting the utility of protein-informed representations for both molecular and morphological prediction. However, its competitive AUC and weaker ACC on cross-cohort binary lung subtyping indicate that transferability remains dependent on the endpoint and evaluation metric.

\subsection{Ablation Studies}

\paragraph{PFE pretraining.}
We evaluate proteomic encoders through masked-protein retrieval on held-out TCGA and external CPTAC profiles, together with their downstream distillation performance (Table~\ref{tab:protein_model_mask_retrieval}). For $\mathrm{M}i$ R@1, we mask $i$ protein measurements in each query profile and compare its embedding against all original profile embeddings in the corresponding gallery. R@1 reports the percentage of queries whose matching original profile ranks first. Distillation results report the mean ACC and AUC across the 12 downstream tasks after using each proteomic encoder as the teacher for ProSlide. PFE provides stronger distillation supervision than Protein-BERT even without virtual profiles, while moderate virtual profile generation improves both retrieval and downstream performance. The largest virtual batch achieves the highest downstream AUC despite weaker retrieval, indicating that retrieval accuracy alone does not fully characterize the value of a proteomic teacher.

\begin{table*}[t]
\centering
\caption{Ablation study of supervision and knowledge distillation.
S2-Sub and S2-Prot denote Stage II subtype supervision and protein-expression supervision, respectively.
S3-KD denotes Stage III knowledge distillation.
Eff. rank denotes the effective rank of representations on the CPTAC datasets.}
\label{tab:stage_ablation}
\setlength{\tabcolsep}{5pt}
\begin{tabular}{cccccccccc}
\toprule
\multirow{2}{*}{S2-Sub}
& \multirow{2}{*}{S2-Prot}
& \multirow{2}{*}{S3-KD}
& \multirow{2}{*}{Eff. rank}
& \multicolumn{2}{c}{Renal}
& \multicolumn{2}{c}{Breast}
& \multicolumn{2}{c}{Lung} \\
\cmidrule(lr){5-6}
\cmidrule(lr){7-8}
\cmidrule(lr){9-10}
& & & & ACC & AUC & ACC & AUC & ACC & AUC \\
\midrule
\ding{51} & \ding{55} & \ding{55} & 11.95& 52.61& 69.79& 57.94& 69.56& 60.12& 73.18 \\
\ding{55} & \ding{51} & \ding{55} & 232.92& 58.07& 75.86&69.22 &76.69 &61.34 &75.39 \\
\ding{55} & \ding{55} & \ding{51} &295.43 & 59.44&74.88 &68.20 &77.51 & 61.92&76.94 \\
\ding{51} & \ding{55} & \ding{51} & 12.12& 52.32&70.26 &58.34 &72.48 &61.11 & 74.41\\
\ding{55} & \ding{51} & \ding{51} & 201.29 &\textbf{60.85}
& \textbf{76.67}
& \textbf{70.02}
& \textbf{78.53}
& \textbf{63.17}
& \textbf{77.90}\\
\bottomrule
\end{tabular}
\end{table*}
\begin{table*}[t]
\centering
\caption{\textbf{Ablation of Prot2Path components.} We evaluate relational targets, standalone references, and sub-bank construction.}
\label{tab:prot2path_ablation}
\setlength{\tabcolsep}{5pt}
\begin{tabular}{lllcccccc}
\toprule
\multirow{2}{*}{Method}&
\multirow{2}{*}{Target}&
\multirow{2}{*}{Reference}

& \multicolumn{2}{c}{Renal}
& \multicolumn{2}{c}{Breast}
& \multicolumn{2}{c}{Lung} \\
\cmidrule(lr){4-5}
\cmidrule(lr){6-7}
\cmidrule(lr){8-9}
 & & & ACC & AUC & ACC & AUC & ACC & AUC \\
\midrule
Direct alignment       & Cosine    & Paired only      &60.07 & \textbf{77.86}& 69.22& \textbf{78.69}&62.34 & 77.39 \\
Hard matching     & CE & All profiles & 60.83& 76.49 & 68.47 & 76.57  &62.89 & 77.69  \\
Paired-bank KD & Soft KL & Paired only         & 60.62 & 76.61 & 69.35 & 77.65 & 62.69 & 77.47 \\
Full-bank KD  & Soft KL & All profiles & 60.33 & 76.90 & \textbf{70.14} & \textbf{78.69} & 62.60 &77.77   \\
Prot2Path              & Soft KL   & Sub-bank  &\textbf{60.85}
& {76.67}
& {70.02}
& {78.53}
& \textbf{63.17}
& \textbf{77.90}\\
\bottomrule
\end{tabular}
\end{table*}
\paragraph{Supervision and training stages.}

To assess the roles of Stage II supervision and Stage III distillation, we compare supervision strategies and combinations of the two stages (Table~\ref{tab:stage_ablation}). Subtype-supervised pretraining produces low-effective-rank representations that remain low-rank after distillation and yield weaker downstream performance. Protein-expression supervision instead produces higher-rank representations and a more effective initialization for knowledge transfer. Combining protein-supervised pretraining with Prot2Path achieves the best ACC and AUC across all three cancer groups, outperforming either stage alone and supporting their complementary contributions.

\paragraph{Prot2Path components.}
To examine the contributions of relational supervision and reference-bank construction, we compare alternative alignment objectives and reference configurations (Table~\ref{tab:prot2path_ablation}). Prot2Path outperforms hard matching and paired-bank KD in both ACC and AUC across all three cancer groups, supporting soft relational targets and the inclusion of standalone molecular references. Compared with direct alignment, its gains are consistent in ACC, while AUC favors different methods across cancer groups. Sampling standalone references maintains competitive performance with fewer reference entries, although neither sampled nor full-bank distillation consistently dominates across metrics. These findings support the effectiveness of the combined design while identifying reference sampling as a trade-off between bank size and predictive performance.

\subsection{Interpretability Analysis}
\begin{figure}[t]
\begin{center}
\includegraphics[width=\linewidth]{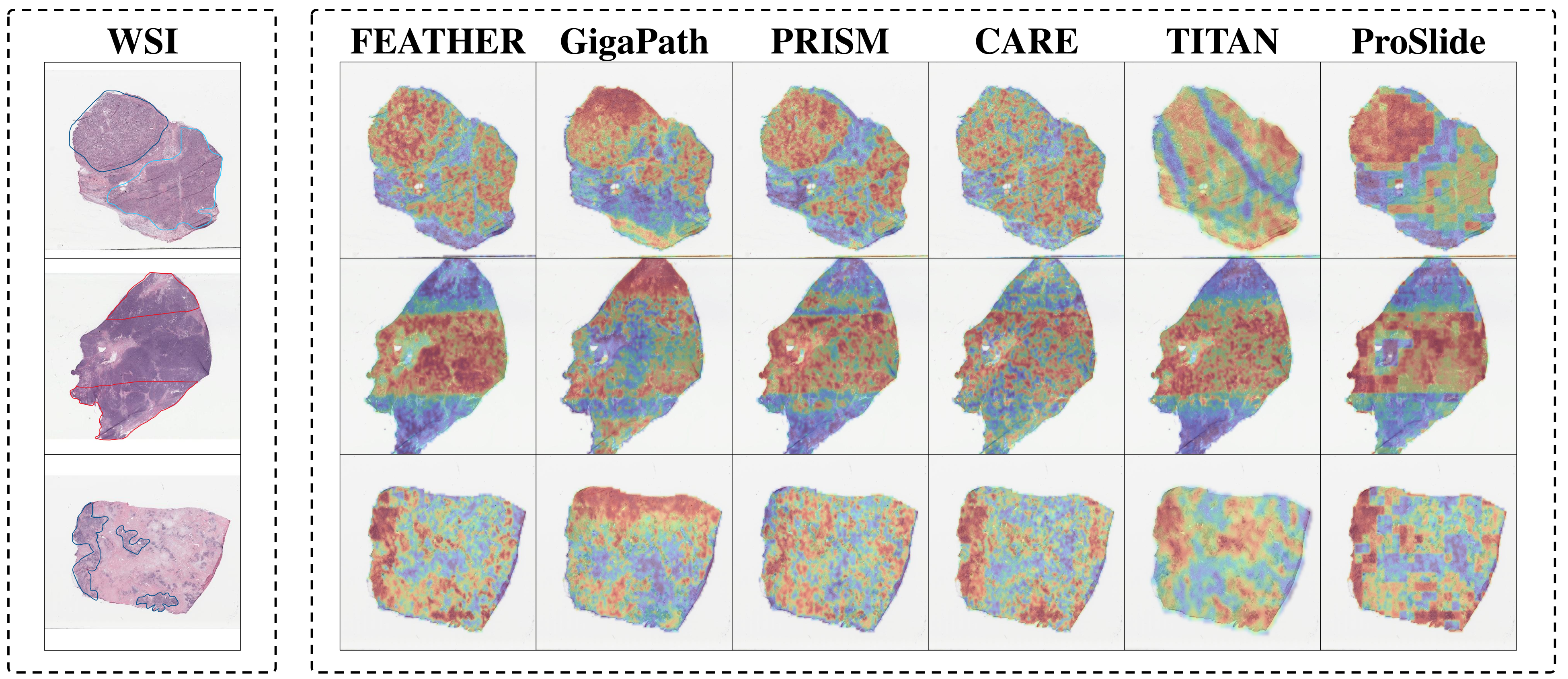}
\end{center}
\caption{Visualization comparison of original WSIs and heatmaps. Heatmaps overlay patch aggregation weights for multiple-instance models or final-layer [CLS]-to-image-token attention weights for transformer-based models onto the corresponding WSIs.}
\label{vis}
\end{figure}

To examine the pathological relevance of ProSlide's attention, we compare its heatmaps with those of other foundation models on three representative CPTAC-BRCA cases (Fig.~\ref{vis}). The annotations provided by an invited pathologist support the assessment of tissue composition, preparation artifacts, and tumor localization.
In the first case, tumor occupies most of the tissue, but the light-blue region contains more inflammatory and stromal components than the dark-blue region. ProSlide assigns stronger attention to the dark-blue region, indicating preferential attention to relatively tumor-rich tissue. In the second case, the tissue is also predominantly tumor-containing, while the red contours identify blurred areas introduced during slide preparation. ProSlide assigns lower attention to these areas and emphasizes the better-preserved central tissue.
In the third case, the blue contours delineate localized tumor regions. ProSlide's higher-attention areas broadly overlap with the annotated regions along the left and lower portions of the tissue. Together, these examples suggest that ProSlide's attention is sensitive to tumor distribution, tissue composition, and image quality, providing complementary qualitative evidence for the pathological relevance of its representations.

\section{Conclusion}
This work investigated whether protein-derived knowledge could improve pathology foundation models. We introduced a three-stage framework that combined independent PFE pretraining, protein-supervised ProSlide pretraining, and Prot2Path relational distillation. Standalone protein profiles contributed to both proteomic representation learning and the distillation reference bank, extending molecular supervision beyond paired cohorts. Evaluation across 12 tasks in breast, renal, and lung cancers supported the effectiveness of this approach and highlighted the value of standalone proteomics for data-efficient pathology representation learning.






\bibliography{references}

@article{pardoll2012blockade,
  title={The blockade of immune checkpoints in cancer immunotherapy},
  author={Pardoll, Drew M},
  journal={Nature reviews cancer},
  volume={12},
  number={4},
  pages={252--264},
  year={2012},
  publisher={Nature Publishing Group UK London}
}

@article{gao2026alpaca,
  title={ALPaCA: Adapting Llama for Pathology Context Analysis to enable slide-level question answering},
  author={Gao, Zeyu and He, Kai and Su, Weiheng and Pang, Xiaobo and Machado, Ines P and Jimenez-Linan, Mercedes and Rous, Brian and Wang, Chunbao and Li, Chengzu and McGough, William and others},
  journal={Nature Communications},
  year={2026},
  publisher={Nature Publishing Group UK London}
}

@article{gao2025smmile,
  title={SMMILe enables accurate spatial quantification in digital pathology using multiple-instance learning},
  author={Gao, Zeyu and Mao, Anyu and Dong, Yuxing and Clayton, Hannah and Wu, Jialun and Liu, Jiashuai and Wang, ChunBao and He, Kai and Gong, Tieliang and Li, Chen and others},
  journal={Nature cancer},
  volume={6},
  number={12},
  pages={2025--2041},
  year={2025},
  publisher={Nature Publishing Group US New York}
}

@article{ahlskog2016atm,
  title={ATM/ATR-mediated phosphorylation of PALB2 promotes RAD51 function},
  author={Ahlskog, Johanna K and Larsen, Brian D and Achanta, Kavya and S{\o}rensen, Claus S},
  journal={The EMBO Reports},
  volume={17},
  number={5},
  pages={671--681},
  year={2016},
  publisher={Springer}
}

@inproceedings{zhang2026care,
	title={CARE: A Molecular-Guided Foundation Model with Adaptive Region Modeling for Whole Slide Image Analysis},
	author={Zhang, Di and Gong, Zhangpeng and Pang, Xiaobo and Liu, Jiashuai and Lu, Junbo and Cui, Hao and Ge, Jiusong and Zeng, Zhi and Yi, Kai and Li, Yinghua and others},
	booktitle={Proceedings of the IEEE/CVF Conference on Computer Vision and Pattern Recognition},
	pages={21078--21088},
	year={2026}
}

@inproceedings{li2025mico,
  title={{MiCo}: Multiple Instance Learning with Context-Aware Clustering for Whole Slide Image Analysis},
  author={Li, Junjian and Liu, Jin and Kuang, Hulin and Yue, Hailin and He, Mengshen and Wang, Jianxin},
  booktitle={International Conference on Medical Image Computing and Computer-Assisted Intervention},
  pages={376--385},
  year={2025},
  organization={Springer}
}

@article{li20252,
  title={{CA2CL}: Cluster-Aware Adversarial Contrastive Learning for Pathological Image Analysis},
  author={Li, Junjian and Kuang, Hulin and Liu, Jin and Yue, Hailin and Wang, Jianxin},
  journal={IEEE Journal of Biomedical and Health Informatics},
  year={2025},
  volume={29},
  number={7},
  pages={5095-5108},
  publisher={IEEE}
}

@article{acosta2022intratumoral,
	title={Intratumoral resolution of driver gene mutation heterogeneity in renal cancer using deep learning},
	author={Acosta, Paul H and Panwar, Vandana and Jarmale, Vipul and Christie, Alana and Jasti, Jay and Margulis, Vitaly and Rakheja, Dinesh and Cheville, John and Leibovich, Bradley C and Parker, Alexander and others},
	journal={Cancer Research},
	volume={82},
	number={15},
	pages={2792--2806},
	year={2022},
	publisher={American Association for Cancer Research}
}

@article{xu2021predicting,
	title={Predicting axillary lymph node metastasis in early breast cancer using deep learning on primary tumor biopsy slides},
	author={Xu, Feng and Zhu, Chuang and Tang, Wenqi and Wang, Ying and Zhang, Yu and Li, Jie and Jiang, Hongchuan and Shi, Zhongyue and Liu, Jun and Jin, Mulan},
	journal={Frontiers in Oncology},
	volume={11},
	pages={759007},
	year={2021},
	publisher={Frontiers Media SA}
}

@article{wei2019pathologist,
	title={Pathologist-level classification of histologic patterns on resected lung adenocarcinoma slides with deep neural networks},
	author={Wei, Jason W and Tafe, Laura J and Linnik, Yevgeniy A and Vaickus, Louis J and Tomita, Naofumi and Hassanpour, Saeed},
	journal={Scientific reports},
	volume={9},
	number={1},
	pages={3358},
	year={2019},
	publisher={Nature Publishing Group UK London}
}

@article{gigapath,
	title={A whole-slide foundation model for digital pathology from real-world data},
	author={Xu, Hanwen and Usuyama, Naoto and Bagga, Jaspreet and Zhang, Sheng and Rao, Rajesh and Naumann, Tristan and Wong, Cliff and Gero, Zelalem and Gonz{\'a}lez, Javier and Gu, Yu and others},
	journal={Nature},
	volume={630},
	number={8015},
	pages={181--188},
	year={2024},
	publisher={Nature Publishing Group UK London}
}

@InProceedings{feather,
  title = 	 {Do Multiple Instance Learning Models Transfer?},
  author =       {Shao, Daniel and Chen, Richard J. and Song, Andrew H. and Runevic, Joel and Lu, Ming Y. and Ding, Tong and Mahmood, Faisal},
  booktitle = 	 {Proceedings of the 42nd International Conference on Machine Learning},
  pages = 	 {54219--54238},
  year = 	 {2025},
  volume = 	 {267},
  series = 	 {Proceedings of Machine Learning Research},
  publisher =    {PMLR},
}

@article{uni, 
	title={Towards a general-purpose foundation model for computational pathology},
	author={Chen, Richard J and Ding, Tong and Lu, Ming Y and Williamson, Drew FK and Jaume, Guillaume and Song, Andrew H and Chen, Bowen and Zhang, Andrew and Shao, Daniel and Shaban, Muhammad and others},
	journal={Nature Medicine},
	volume={30},
	number={3},
	pages={850--862},
	year={2024},
	publisher={Nature Publishing Group US New York}
}

@article{conch,
  title={A visual-language foundation model for computational pathology},
  author={Lu, Ming Y and Chen, Bowen and Williamson, Drew FK and Chen, Richard J and Liang, Ivy and Ding, Tong and Jaume, Guillaume and Odintsov, Igor and Le, Long Phi and Gerber, Georg and others},
  journal={Nature Medicine},
  volume={30},
  number={3},
  pages={863--874},
  year={2024},
  publisher={Nature Publishing Group US New York}
}

@article{esm,
	title={Language models of protein sequences at the scale of evolution enable accurate structure prediction},
	author={Lin, Zeming and Akin, Halil and Rao, Roshan and Hie, Brian and Zhu, Zhongkai and Lu, Wenting and Smetanin, Nikita and dos Santos Costa, Allan and Fazel-Zarandi, Maryam and Sercu, Tom and Candido, Sal and others},
	journal={bioRxiv},
	year={2022},
	publisher={Cold Spring Harbor Laboratory}
}

@article{cui2023scGPT,
	title={scGPT: Towards Building a Foundation Model for Single-Cell Multi-omics Using Generative AI},
	author={Cui, Haotian and Wang, Chloe and Maan, Hassaan and Pang, Kuan and Luo, Fengning and Wang, Bo},
	journal={bioRxiv},
	year={2023},
	publisher={Cold Spring Harbor Laboratory}
}

@inproceedings{hipt,
	title={Scaling vision transformers to gigapixel images via hierarchical self-supervised learning},
	author={Chen, Richard J and Chen, Chengkuan and Li, Yicong and Chen, Tiffany Y and Trister, Andrew D and Krishnan, Rahul G and Mahmood, Faisal},
	booktitle={Proceedings of the IEEE/CVF Conference on Computer Vision and Pattern Recognition},
	pages={16144--16155},
	year={2022}
}

@article{prism,
	title={Prism: A multi-modal generative foundation model for slide-level histopathology},
	author={Shaikovski, George and Casson, Adam and Severson, Kristen and Zimmermann, Eric and Wang, Yi Kan and Kunz, Jeremy D and Retamero, Juan A and Oakley, Gerard and Klimstra, David and Kanan, Christopher and others},
	journal={arXiv preprint arXiv:2405.10254},
	year={2024}
}

@article{yang2026path,
      title={PathNavigate: A Training-Free Pathology Agent with Surprise-Guided Scan and Shared Slide Memory for Whole-Slide Image VQA}, 
      author={Chunze Yang and Qidong Liu and Wenjie Zhao and Yue Tang and Jiusong Ge and Di Zhang and Jiashuai Liu and Lei Wu and Junbo Lu and Ni Zhang and Xian Wu and Zeyu Gao and Chen Li},
      year={2026},
      journal={arXiv preprint arXiv:2605.23559},
}

@article{titan,
	title={A multimodal whole-slide foundation model for pathology},
	author={Ding, Tong and Wagner, Sophia J and Song, Andrew H and Chen, Richard J and Lu, Ming Y and Zhang, Andrew and Vaidya, Anurag J and Jaume, Guillaume and Shaban, Muhammad and Kim, Ahrong and others},
	journal={Nature Medicine},
	volume={31},
	number={11},
	pages={3749--3761},
	year={2025}
}

@article{li2024scprotein,
  title={scPROTEIN: a versatile deep graph contrastive learning framework for single-cell proteomics embedding},
  author={Li, Wei and Yang, Fan and Wang, Fang and Rong, Yu and Liu, Linjing and Wu, Bingzhe and Zhang, Han and Yao, Jianhua},
  journal={Nature Methods},
  volume={21},
  number={4},
  pages={623--634},
  year={2024},
  publisher={Nature Publishing Group US New York}
}

@inproceedings{guo2026momentum,
  title={Momentum memory for knowledge distillation in computational pathology},
  author={Guo, Yongxin and Lu, Hao and Koyun, Onur C and Zhu, Zhengjie and Demir, Muhammet F and Gurcan, Metin N},
  booktitle={Proceedings of the IEEE/CVF Conference on Computer Vision and Pattern Recognition},
  pages={6889--6899},
  year={2026}
}

@article{jumper2021highly,
  title={Highly accurate protein structure prediction with AlphaFold},
  author={Jumper, John and Evans, Richard and Pritzel, Alexander and Green, Tim and Figurnov, Michael and Ronneberger, Olaf and Tunyasuvunakool, Kathryn and Bates, Russ and {\v{Z}}{\'\i}dek, Augustin and Potapenko, Anna and others},
  journal={nature},
  volume={596},
  number={7873},
  pages={583--589},
  year={2021},
  publisher={Nature Publishing Group UK London}
}

@article{abramson2024accurate,
  title={Accurate structure prediction of biomolecular interactions with AlphaFold 3},
  author={Abramson, Josh and Adler, Jonas and Dunger, Jack and Evans, Richard and Green, Tim and Pritzel, Alexander and Ronneberger, Olaf and Willmore, Lindsay and Ballard, Andrew J and Bambrick, Joshua and others},
  journal={Nature},
  volume={630},
  number={8016},
  pages={493--500},
  year={2024},
  publisher={Nature Publishing Group UK London}
}

@article{zhou2026scpformer,
  title={scpFormer: A Foundation Model for Unified Representation and Integration of the Single-Cell Proteomics},
  author={Zhou, Qifeng and Yu, Lei and Guo, Yuzhi and Miao, Yuwei and Ma, Hehuan and Zhong, Wenliang and Xu, Lin and Huang, Junzhou},
  journal={arXiv preprint arXiv:2604.20003},
  year={2026}
}

@article{wenckstern2026virtual,
  title={The Virtual Tissues foundation model resolves spatial proteomics across scales},
  author={Wenckstern, Johann and Jain, Eeshaan and von Querfurth, Benedikt and Cheng, Yexiang and Vasilev, Kiril and Pariset, Matteo and Cheng, Phil F and Liakopoulos, Petros and Michielin, Olivier and Wicki, Andreas and others},
  journal={Nature},
  pages={1--11},
  year={2026},
  publisher={Nature Publishing Group UK London}
}

@inproceedings{park2019relational,
  title={Relational knowledge distillation},
  author={Park, Wonpyo and Kim, Dongju and Lu, Yan and Cho, Minsu},
  booktitle={2019 IEEE/CVF Conference on Computer Vision and Pattern Recognition (CVPR)},
  pages={3962--3971},
  year={2019},
  organization={IEEE}
}

@article{threads,
  title={Molecular-driven foundation model for oncologic pathology},
  author={Vaidya, Anurag and Zhang, Andrew and Jaume, Guillaume and Song, Andrew H and Ding, Tong and Wagner, Sophia J and Lu, Ming Y and Doucet, Paul and Robertson, Harry and Almagro-Perez, Cristina and others},
  journal={arXiv preprint arXiv:2501.16652},
  year={2025}
}

@article{xu2025multimodal,
  title={A multimodal knowledge-enhanced whole-slide pathology foundation model},
  author={Xu, Yingxue and Wang, Yihui and Zhou, Fengtao and Ma, Jiabo and Jin, Cheng and Yang, Shu and Li, Jinbang and Zhang, Zhengyu and Zhao, Chenglong and Zhou, Huajun and others},
  journal={Nature communications},
  volume={16},
  number={1},
  pages={11406},
  year={2025},
  publisher={Nature Publishing Group UK London}
}

@article{lffa,
  title={Learnable frozen feature augmentation for few-shot biomarker prediction from pathology whole-slide images},
  author={Zhang, Di and Liu, Jiashuai and Ma, Youyuan and Ge, Jiusong and Zeng, Zhi and Sun, Wenfang and Liu, Qidong and He, Kai and Zheng, Yefeng and Yu, Weimiao and others},
  journal={Bioinformatics},
  pages={btag597},
  year={2026},
  publisher={Oxford University Press}
}

@article{dinov2,
  title={Dinov2: Learning robust visual features without supervision},
  author={Oquab, Maxime and Darcet, Timoth{\'e}e and Moutakanni, Th{\'e}o and Vo, Huy and Szafraniec, Marc and Khalidov, Vasil and Fernandez, Pierre and Haziza, Daniel and Massa, Francisco and El-Nouby, Alaaeldin and others},
  journal={arXiv preprint arXiv:2304.07193},
  year={2023}
}

@InProceedings{grashei2025pathryoshka,
author="Grashei, Christian
and Brechenmacher, Christian
and Umer, Rao Muhammad
and Liu, Jingsong
and Marr, Carsten
and Sch{\"u}ffler, Peter
and Szczurek, Ewa",
title="Pathryoshka: Compressing pathology foundation models via multi-teacher knowledge distillation with nested embeddings",
booktitle="European Conference on Computer Vision",
year="2026",
publisher="Springer Nature Switzerland",
address="Cham",
pages="76--94",
isbn="978-3-032-37029-7"
}

@article{akbani2014pan,
  title={A pan-cancer proteomic perspective on The Cancer Genome Atlas},
  author={Akbani, Rehan and Ng, Patrick Kwok Shing and Werner, Henrica MJ and Shahmoradgoli, Maria and Zhang, Fan and Ju, Zhenlin and Liu, Wenbin and Yang, Ji-Yeon and Yoshihara, Kosuke and Li, Jun and others},
  journal={Nature communications},
  volume={5},
  number={1},
  pages={3887},
  year={2014},
  publisher={Nature Publishing Group UK London}
}

@article{li2013tcpa,
  title={TCPA: a resource for cancer functional proteomics data},
  author={Li, Jun and Lu, Yiling and Akbani, Rehan and Ju, Zhenlin and Roebuck, Paul L and Liu, Wenbin and Yang, Ji-Yeon and Broom, Bradley M and Verhaak, Roeland GW and Kane, David W and others},
  journal={Nature methods},
  volume={10},
  number={11},
  pages={1046--1047},
  year={2013},
  publisher={Nature Publishing Group US New York}
}

@article{ghandi2019next,
  title={Next-generation characterization of the cancer cell line encyclopedia},
  author={Ghandi, Mahmoud and Huang, Franklin W and Jan{\'e}-Valbuena, Judit and Kryukov, Gregory V and Lo, Christopher C and McDonald III, E Robert and Barretina, Jordi and Gelfand, Ellen T and Bielski, Craig M and Li, Haoxin and others},
  journal={Nature},
  volume={569},
  number={7757},
  pages={503--508},
  year={2019},
  publisher={Nature Publishing Group UK London}
}

@article{zhao2020large,
  title={Large-scale characterization of drug responses of clinically relevant proteins in cancer cell lines},
  author={Zhao, Wei and Li, Jun and Chen, Mei-Ju M and Luo, Yikai and Ju, Zhenlin and Nesser, Nicole K and Johnson-Camacho, Katie and Boniface, Christopher T and Lawrence, Yancey and Pande, Nupur T and others},
  journal={Cancer cell},
  volume={38},
  number={6},
  pages={829--843},
  year={2020},
  publisher={Elsevier}
}

@article{edwards2015cptac,
  title={The CPTAC data portal: a resource for cancer proteomics research},
  author={Edwards, Nathan J and Oberti, Mauricio and Thangudu, Ratna R and Cai, Shuang and McGarvey, Peter B and Jacob, Shine and Madhavan, Subha and Ketchum, Karen A},
  journal={Journal of proteome research},
  volume={14},
  number={6},
  pages={2707--2713},
  year={2015},
  publisher={ACS Publications}
}
\bibliographystyle{plainnat}

\clearpage
\appendix
\renewcommand{\topfraction}{0.95}
\renewcommand{\textfraction}{0.05}
\renewcommand{\floatpagefraction}{0.7}
\setcounter{figure}{0}
\setcounter{table}{0}
\renewcommand{\thefigure}{S\arabic{figure}}
\renewcommand{\thetable}{S\arabic{table}}

\ifdefined\theHfigure
  \renewcommand{\theHfigure}{supp.\arabic{figure}}
\fi
\ifdefined\theHtable
  \renewcommand{\theHtable}{supp.\arabic{table}}
\fi
\begin{center}

\vspace{1.6em}
{\Large\bfseries Can Protein-Derived Knowledge \\ Improve Pathology Foundation Models?}\par
\vspace{0.25em}
{\large\bfseries\itshape Supplementary Material}\par
\vspace{1.7em}
\rule{0.94\linewidth}{0.4pt}
\end{center}
\vspace{1.5em}
\noindent{\Large\bfseries Table of Contents}
\vspace{0.2em}
\hrule
\vspace{0.9em}

\begingroup
\small
\makeatletter
\newcommand{\appdirsection}[3]{%
	\par\addvspace{0.95em}%
	\noindent\makebox[2.4em][l]{\bfseries #1}%
	{\bfseries #2}\nobreak\hfill{\bfseries \pageref{#3}}\par}
\newcommand{\appdirsubsection}[3]{%
	\@dottedtocline{1}{2.4em}{3.4em}{#1\quad #2}{\pageref{#3}}}
\makeatother
\appdirsection{A}{Proteomic Data and Preprocessing}{app:protein_data}
\appdirsubsection{A.1}{Proteomic Data Resources}{ProteomicDataResources}
\appdirsubsection{A.2}{Cross-Resource Protein Harmonization}{CrossResourceProteinHarmonization}
\appdirsubsection{A.3}{Missing Measurements and Expression Normalization}{Missing}
\appdirsubsection{A.4}{Selection of the 38-Protein Panel}{Selectionofthe}

\appdirsection{B}{PFE Pretraining}{app:protein_pretrain}
\appdirsubsection{B.1}{Details of Virtual Profile Generation}{app:pretrain_details}
\appdirsubsection{B.2}{Details of Expression-Space Multi-View Construction}{DetailsofExpression}
\appdirsubsection{B.3}{Profile- and Protein-Level Learning Objectives}{supp:loss}
\appdirsubsection{B.4}{PFE Pretraining Configuration}{Configuration}

\appdirsubsection{B.5}{Comparison of Protein Modeling Paradigms}{ProteinModelingParadigms}

\appdirsection{C}{ProSlide Pretraining and Prot2Path Distillation}{app:pathology_pretraining}
\appdirsubsection{C.1}{ProSlide Architecture}{ProSlideArchitecture}
\appdirsubsection{C.2}{Prot2Path Implementation Details}{Prot2PathImplementation}

\appdirsection{D}{Downstream Evaluation and Additional Analyses}{app:downstream}
\appdirsubsection{D.1}{Datasets and Task Definitions}{app:Datasets}
\appdirsubsection{D.2}{Evaluation Protocols and Baselines}{app:Evaluation}
\appdirsubsection{D.3}{Detailed Downstream Results}{app:Results}
\appdirsubsection{D.4}{Data Efficiency and Inference Memory Usage}{app:Efficiency}
\appdirsubsection{D.5}{Analysis of PFE Protein Identity Embeddings}{app:Interpretability}
\endgroup

\vfill
\hrule

\clearpage

\section{Proteomic Data and Preprocessing}
\label{app:protein_data}

To facilitate reproducibility, we provide additional details on the proteomic datasets and the corresponding preprocessing procedures.
All four resources contain reverse-phase protein array (RPPA) measurements, which provide antibody-based quantification of both total proteins and phosphorylation-specific protein targets. After harmonization, all samples are represented using a common panel of 38 cancer-related protein measurements.

\subsection{Proteomic Data Resources}
\label{ProteomicDataResources}
We integrate RPPA profiles from four complementary resources: TCGA, TCPA, CCLE, and CPPA. These resources encompass patient tumor specimens, baseline cancer cell lines, and pharmacologically perturbed cancer cell lines, providing diverse protein-expression patterns for PFE pretraining. As summarized in Table~\ref{tab:pfe_pretrain_data}, the resulting corpus comprises 13,586 protein profiles: 7,349 from TCGA, 878 from TCPA, 899 from CCLE, and 4,460 from CPPA. We curated a total of 13,586 proteomic profiles. Of these, 891 TCGA profiles were held out for evaluation, and the remaining 12,695 profiles were used for PFE pretraining.

\paragraph{The Cancer Genome Atlas (TCGA).}
The Cancer Genome Atlas (TCGA) is a large-scale pan-cancer initiative that provides genomic, molecular, and histopathological characterization across a broad range of human malignancies \citep{akbani2014pan}. We use TCGA RPPA measurements as the primary source of patient-derived tumor proteomic profiles. The current cohort contains 7,349 RPPA records spanning 32 cancer cohorts. Unlike the external cell-line resources described below, TCGA provides naturally occurring tumor protein-expression profiles and constitutes the major patient tissue domain in PFE pretraining.

\paragraph{The Cancer Proteome Atlas (TCPA).}
The Cancer Proteome Atlas (TCPA) provides uniformly processed RPPA-based functional proteomic measurements for cancer samples and cell lines \citep{li2013tcpa}. In our pretraining cohort, we use the CCLE RPPA500 profiles distributed through TCPA rather than treating TCPA as an additional independent TCGA patient cohort. After protein-name harmonization and coverage filtering, 878 cancer cell-line profiles are retained. These samples complement patient-derived TCGA tumors by introducing protein-expression patterns from experimentally controlled cancer cell-line systems.

\paragraph{Cancer Cell Line Encyclopedia (CCLE).}
The Cancer Cell Line Encyclopedia (CCLE) provides systematic molecular characterization of cancer cell lines originating from diverse tissue types \citep{ghandi2019next}. We use the CCLE RPPA measurements together with the corresponding antibody annotations to resolve protein identities and phosphorylation-specific targets. Following harmonization with the common 38-protein panel, 899 baseline cancer cell-line profiles are included in PFE pretraining. In contrast to TCGA tumor specimens, these samples represent protein expression in unperturbed cancer cell lines and therefore provide an additional molecular context for learning reusable proteomic representations.

\paragraph{Cancer Perturbed Proteomics Atlas (CPPA).}
The Cancer Perturbed Proteomics Atlas (CPPA) characterizes protein responses of cancer cell lines under pharmacological perturbations \citep{zhao2020large}. Unlike TCGA, TCPA, and CCLE profiles that primarily describe protein abundance states, CPPA records treatment-induced changes in protein abundance. Specifically, a perturbation response is represented as
$\Delta p = p_{1} - p_{0},$
where $p_{0}$ denotes the baseline protein level before treatment and $p_{1}$ denotes the corresponding level after drug exposure. We retain 4,460 CPPA perturbation records after protein mapping and coverage filtering. These profiles expose PFE to drug-, dose-, and treatment-dependent protein responses, thereby broadening the molecular variation available during pretraining. Importantly, each CPPA record corresponds to a perturbation condition rather than an independent cell line.

\begin{table}[t]
\centering
\caption{Composition of the proteomic cohort.}
\label{tab:pfe_pretrain_data}
\small
\setlength{\tabcolsep}{5pt}
\begin{tabular}{lcll}
\toprule
\textbf{Resource} & \textbf{ Records} & \textbf{Sample Domain} & \textbf{Protein Profile Type} \\
\midrule
TCGA & 7,349 & Patient tumors & Baseline tumor RPPA \\
TCPA & 878 & Cancer cell lines & Baseline cell-line RPPA \\
CCLE & 899 & Cancer cell lines & Baseline cell-line RPPA \\
CPPA & 4,460 & Perturbed cell lines & Treatment-induced $\Delta p$ \\
\midrule
\textbf{Total} & \textbf{13,586} & -- & -- \\
\bottomrule
\end{tabular}
\end{table}

\subsection{Cross-Resource Protein Harmonization}
\label{CrossResourceProteinHarmonization}
The original RPPA resources contain different antibody panels and use heterogeneous naming conventions for proteins and phosphorylation sites. We therefore map all sources to a fixed panel of $M=38$ cancer-related protein targets before PFE pretraining (see Appendix~\ref{Selectionofthe}). Protein names are first standardized by normalizing capitalization and removing formatting differences and technical suffixes. Manually curated aliases are further used to reconcile established synonyms across resources. For phosphorylation-specific measurements, mapping is performed conservatively: only measurements corresponding to the same protein and the same modification site are treated as equivalent, while distinct phosphorylation sites are retained as separate molecular targets.

\subsection{Missing Measurements and Expression Normalization}
\label{Missing}
Because individual RPPA resources do not uniformly measure all 38 target proteins, some protein measurements may be unavailable for a given sample. Missing entries arise when a target is absent from the original antibody panel, cannot be reliably mapped across resources, or contains an invalid measurement. These entries are explicitly marked as missing and are handled by the learnable mask representation described in Eq. \ref{protein_token}.
To maintain the original characteristics of each RPPA resource, we do not impose an additional cross-resource global normalization when constructing the unified dataset. Instead, source measurements are retained on their original processed scales after protein harmonization. During PFE training, valid expression measurements are standardized using protein-specific statistics before being mapped into protein tokens. This separates the preservation of the original RPPA measurements from the numerical normalization required for stable representation learning.
External cell-line and perturbation samples are additionally filtered according to protein coverage. Only profiles containing measurements for a sufficient number of targets in the shared protein panel are retained, preventing extremely sparse profiles from dominating the self-supervised pretraining process.

\subsection{Selection of the 38-Protein Panel}
\label{Selectionofthe}

We use a fixed panel of 38 cancer-related RPPA targets to establish a shared input space across the integrated proteomic resources. The panel spans DNA damage response and repair, growth signaling, cell-cycle and chromatin regulation, hormone receptor signaling, lineage-associated transcription, immune regulation, and hypoxia-related processes. This composition provides complementary molecular measurements relevant to tumor biology and the cellular context of bulk specimens. Table~\ref{tab:protein_panel} lists the panel entries, standardized targets, and functional annotations.
The panel comprises 32 distinct proteins and six additional site-specific phosphorylation measurements involving ESR1, ERBB2, EGFR, MET, and MTOR. Each phosphorylation measurement is retained as a separate target from its corresponding total-protein measurement, preserving the distinction between protein abundance and site-specific phosphorylation. 

\begin{table}[!t]
	\centering
	\caption{\textbf{Composition of the 38-protein panel.} Total-protein and site-specific phosphorylation measurements are retained as distinct targets.}
	\label{tab:protein_panel}
	\small
	\setlength{\tabcolsep}{5pt}
	\renewcommand{\arraystretch}{1.05}
	\begin{tabularx}{\linewidth}{@{}llX@{}}
		\toprule
		\textbf{Panel Name} & \textbf{Standardized Target} & \textbf{Functional Annotation} \\
		\midrule
		ATM & ATM & DNA damage response \\
		ATR & ATR & DNA damage response \\
		BRCA2 & BRCA2 & Homologous recombination repair \\
		P53 & TP53 & Tumor suppression / DNA damage response \\
		RB & RB1 & Cell-cycle regulation \\
		BAP1C4 & BAP1 & Chromatin regulation / tumor suppression \\
		SETD2 & SETD2 & Chromatin regulation / tumor suppression \\
		AXL & AXL & Receptor tyrosine kinase signaling \\
		BRAF & BRAF & RAS--MAPK signaling \\
		NRAS & NRAS & RAS--MAPK signaling \\
		EGFR & EGFR & Receptor tyrosine kinase signaling \\
		EGFR\_pY1068 & EGFR pTyr1068 & Receptor tyrosine kinase signaling \\
		EGFR\_pY1173 & EGFR pTyr1173 & Receptor tyrosine kinase signaling \\
		HER2 & ERBB2 & Receptor tyrosine kinase signaling \\
		HER2\_pY1248 & ERBB2 pTyr1248 & Receptor tyrosine kinase signaling \\
		CMET & MET & Receptor tyrosine kinase signaling \\
		CMET\_pY1235 & MET pTyr1235 & Receptor tyrosine kinase signaling \\
		PDGFRB & PDGFRB & Receptor tyrosine kinase signaling \\
		MTOR & MTOR & PI3K--AKT--mTOR signaling \\
		MTOR\_pS2448 & MTOR pSer2448 & PI3K--AKT--mTOR signaling \\
		PTEN & PTEN & Negative regulation of PI3K--AKT signaling \\
		SMAD4 & SMAD4 & TGF-$\beta$ signaling \\
		ERALPHA & ESR1 (ER-$\alpha$) & Hormone receptor signaling \\
		ERALPHA\_pS118 & ESR1 pSer118 & Hormone receptor signaling \\
		PR & PGR & Hormone receptor signaling \\
		GATA3 & GATA3 & Lineage-associated transcriptional regulation \\
		MITF & MITF & Lineage-associated transcriptional regulation \\
		PAX8 & PAX8 & Lineage-associated transcriptional regulation \\
		
		CD4 & CD4 & Immune-cell marker / T-cell co-receptor \\
		CD86 & CD86 & Immune co-stimulation \\
		CIITA & CIITA & Regulation of antigen presentation \\
		CTLA4 & CTLA4 & Immune checkpoint \\
		PDCD1 & PDCD1 (PD-1) & Immune checkpoint \\
		PDL1 & CD274 (PD-L1) & Immune checkpoint ligand \\
		CD44 & CD44 & Cell adhesion / stemness-associated processes \\
		MYH11 & MYH11 & Smooth-muscle contractile machinery \\
		VHL & VHL & Regulation of hypoxia signaling \\
		CA9 & CA9 (CAIX) & Hypoxia-associated pH regulation \\
		\bottomrule
	\end{tabularx}
\end{table}

\clearpage

\section{Proteomic Pretraining }
\label{app:protein_pretrain}

Conventional multi-view self-supervised learning is largely developed for images \citep{dinov2}, where semantically related views can be constructed through spatial cropping and photometric transformations. Directly transferring this strategy to bulk proteomic profiles is non-trivial, because a protein-expression profile has no intrinsic spatial structure, each dimension is tied to an explicitly defined protein identity, and the corresponding measurements are continuous expression values. We therefore reformulate multi-view learning directly in the protein-expression space, preserving protein identities while perturbing only their expression states.

Our expression-space pretraining strategy addresses two complementary challenges of bulk proteomic foundation modeling. First, the limited number of available patient-level protein profiles restricts the diversity of molecular states observed during pretraining. We therefore introduce \emph{Virtual Profile Generation} to expand the effective proteomic cohort from real parent profiles. Second, because bulk protein profiles lack the spatial structure that enables conventional image-based multi-view augmentation, we construct semantically related views directly in the protein-expression space using \emph{profile-wide views} and \emph{subset-perturbed views}. These designs are subsequently coupled with profile- and protein-level self-supervised objectives to learn both global molecular representations and contextualized protein representations.

\subsection{Details of Virtual Profile Generation}
\label{app:pretrain_details}

To clarify how virtual protein profiles are generated, we describe the source-profile sampling procedure and the three expression perturbation operators used during PFE pretraining.

Each mini-batch contains $B_{\mathrm{r}}$ real profiles sampled from $\mathcal{D}$ and $B_{\mathrm{v}}$ virtual profiles generated online. For each virtual profile $a$, we independently sample a source profile $\mathcal{X}_{k_a}$ from the full pretraining corpus. Source selection is not restricted to the real profiles present in the current mini-batch.

Let $\mathcal{O}_{k_a}$ denote the observed proteins in the source profile:
\begin{equation}
\mathcal{O}_{k_a} = \{j \in \{1,\ldots,M\}:m_{k_a,j}=1\}.
\end{equation}
We sample a small subset $\mathcal{S}_a\subseteq\mathcal{O}_{k_a}$ and modify only its expression values according to Eq.~\ref{eq:virtual_profile}. One augmentation type is sampled per virtual profile, with probabilities $\pi_{\mathrm{post}}$, $\pi_{\mathrm{gau}}$, and $\pi_{\mathrm{mix}}$:
\begin{equation}
\pi_{\mathrm{post}}+\pi_{\mathrm{gau}}+\pi_{\mathrm{mix}}=1.
\end{equation}

Next, we introduce three Virtual Profile Generation Operators.

\paragraph{Partial posterization.}
Posterization reduces the numerical precision of selected expression values. Let $\ell_j<u_j$ denote the scaling bounds for protein $j$, and let $K\geq2$ denote the number of quantization levels. For an expression value within these bounds, we first map it to the unit interval, quantize it, and restore the original scale:
\begin{align}
z_{a,j} &= \frac{x_{k_a,j}-\ell_j}{u_j-\ell_j},\\
\bar{z}_{a,j} &= Q_K(z_{a,j}) = \frac{\operatorname{round}\!\left((K-1)z_{a,j}\right)}{K-1},\\
\tilde{x}_{a,j} &= \ell_j+(u_j-\ell_j)\bar{z}_{a,j},
\qquad j\in\mathcal{S}_a.
\label{eq:app_posterization}
\end{align}
The quantizer maps values in $[0,1]$ to $K$ uniformly spaced levels. Inverse scaling restores the expression scale while retaining the loss of within-bin precision introduced by quantization.

\paragraph{Partial Gaussian perturbation.}
For each selected protein, we independently add Gaussian noise on the normalized expression scale:
\begin{equation}
\tilde{x}_{a,j} = x_{k_a,j}+\epsilon_{a,j},
\qquad
\epsilon_{a,j}\overset{\mathrm{i.i.d.}}{\sim}\mathcal{N}(0,\sigma_{\mathrm{virt}}^2),
\qquad j\in\mathcal{S}_a.
\label{eq:app_gaussian}
\end{equation}
The perturbation changes only selected expression values and does not alter protein identities or missingness indicators.

\paragraph{Partial mixup.}
For each virtual profile assigned mixup, we sample a donor uniformly from the corpus, excluding its source:
\begin{equation}
b_a \sim \operatorname{Unif}\!\left(\{1,\ldots,N\}\setminus\{k_a\}\right).
\end{equation}
Interpolation is restricted to selected proteins observed in both profiles:
\begin{equation}
\mathcal{S}^{\mathrm{mix}}_a = \{j\in\mathcal{S}_a:m_{b_a,j}=1\}.
\end{equation}
We draw one coefficient per virtual profile,
\begin{equation}
\lambda_a\sim\operatorname{Beta}(\alpha,\alpha),
\qquad \alpha=0.4,
\end{equation}
and construct the expression values as
\begin{equation}
\tilde{x}_{a,j} =
\begin{cases}
\lambda_a x_{k_a,j}+(1-\lambda_a)x_{b_a,j}, & j\in\mathcal{S}^{\mathrm{mix}}_a,\\
x_{k_a,j}, & \text{otherwise}.
\end{cases}
\label{eq:app_partial_mixup}
\end{equation}
The same $\lambda_a$ is shared across all interpolated proteins. Selected proteins missing from the donor remain unchanged, and the virtual profile retains the source protein identities and missingness pattern.

\subsection{Details of Expression-Space Multi-View Construction}
\label{DetailsofExpression}
To encourage consistent proteomic representations under expression perturbations, we construct complementary profile-wide and subset-perturbed views for each input profile. View construction is applied after real and virtual profiles have been assembled into a mini-batch. Let $\mathcal{X}=\{(p_j,x_j,m_j)\}_{j=1}^{M}$ denote either a real or a virtual profile, with observed set $\mathcal{O}=\{j:m_j=1\}$. All views retain the complete set of $M$ protein identities, with transformations restricted to observed expression values.

\paragraph{Profile-wide views.}
We construct three profile-wide views: one clean view and two stochastically augmented views. The clean view preserves the input profile. Each augmented view applies mild posterization to all observed expression values using independently sampled augmentation settings:
\begin{equation}
x^{(v)}_j =
\begin{cases}
\mathcal{P}_{\theta_v,j}(x_j), & j\in\mathcal{O},\\
x_j, & \text{otherwise},
\end{cases}
\qquad v\in\{1,2\}.
\label{eq:app_profile_wide_views}
\end{equation}
Here, $\mathcal{P}_{\theta_v,j}$ denotes the posterization-based transformation for protein $j$ under the sampled configuration $\theta_v$. Equation~\ref{eq:app_posterization} defines its base quantization operation. Both augmented views preserve protein identities and naturally missing measurements.

\paragraph{Subset-perturbed views.}
We additionally construct $V_{\mathrm{sub}}$ views, each using an independently sampled subset $\mathcal{S}^{(v)}\subseteq\mathcal{O}$. Let $\mathcal{G}^{(v)}\subseteq\mathcal{S}^{(v)}$ and $\mathcal{R}^{(v)}\subseteq\mathcal{S}^{(v)}$ denote the positions selected for Gaussian perturbation and artificial masking, respectively. Before masking, the expression values are
\begin{equation}
x^{(v)}_j =
\begin{cases}
x_j+\epsilon^{(v)}_j, & j\in\mathcal{G}^{(v)},\\
x_j, & \text{otherwise},
\end{cases}
\qquad
\epsilon^{(v)}_j\overset{\mathrm{i.i.d.}}{\sim}\mathcal{N}(0,\sigma_{\mathrm{sub}}^2).
\label{eq:app_subset_noise}
\end{equation}
The expression state at each position in $\mathcal{R}^{(v)}$ is then hidden from the encoder while its protein identity token is retained. Artificially masked positions originate from observed measurements and are distinguished from naturally missing proteins. Positions outside the selected subset remain unchanged.

\begin{table}[t]
\centering
\caption{\textbf{Architecture configuration of the Proteomic Foundation Encoder (PFE).}
PFE represents each protein as the sum of a learnable identity embedding and a continuous abundance-value embedding, followed by Transformer-based contextual modeling.}
\label{tab:pfe_architecture}
\small
\setlength{\tabcolsep}{5pt}
\begin{tabular}{lll}
\toprule
\textbf{Component} & \textbf{Configuration} & \textbf{Description} \\
\midrule
Protein panel size
& 38
& Shared protein vocabulary \\

Input sequence length
& $39$
& One \texttt{[CLS]} and 38 protein tokens \\

Value projection $g_{\mathrm{val}}$
& $1 \rightarrow 32 \rightarrow 768$
& Projects continuous abundance values \\

Missing-value embedding
& $1 \times 768$
& Shared learnable value-mask embedding \\

Hidden dimension
& 768
& Transformer token dimension \\

Transformer depth
& 7
& Number of encoder blocks \\

Attention heads
& 8
& Multi-head self-attention \\

MLP ratio
& 4
& Transformer feed-forward expansion ratio \\

Dropout
& 0.1
& Applied in Transformer blocks \\

Attention dropout
& 0.1
& Applied to self-attention \\

Profile representation
& 768-d \texttt{[CLS]}
& Patient-level molecular representation \\

Protein representation
& 768-d per token
& Contextualized protein-specific representation \\

Projection head
& $768 \rightarrow 2048 \rightarrow 2048 \rightarrow 256$
& DINO-style projection MLP \\

Prototype dimension
& 2048
& Profile- and protein-level soft targets \\

\bottomrule
\end{tabular}
\end{table}

\paragraph{Asymmetric view assignment.}
Let $\mathcal{V}_{\mathrm{wide}}(\mathcal{X})$ and $\mathcal{V}_{\mathrm{sub}}(\mathcal{X})$ denote the profile-wide and subset-perturbed view collections. The teacher and student receive
\begin{equation}
\mathcal{V}_{\mathrm{teacher}}(\mathcal{X}) = \mathcal{V}_{\mathrm{wide}}(\mathcal{X}),
\qquad
\mathcal{V}_{\mathrm{student}}(\mathcal{X}) = \mathcal{V}_{\mathrm{wide}}(\mathcal{X})\cup\mathcal{V}_{\mathrm{sub}}(\mathcal{X}).
\label{eq:app_view_assignment}
\end{equation}
Thus, the teacher processes three profile-wide views, while the student additionally processes $V_{\mathrm{sub}}$ locally perturbed views. This assignment is applied to both real and virtual profiles. The profile- and protein-level learning objectives are defined in Appendix \ref{supp:loss}.

\subsection{Profile- and Protein-Level Learning Objectives}
\label{supp:loss}

To learn proteomic representations at both the profile and protein levels, we pretrain PFE using a student--teacher self-distillation framework. The student parameters $\theta_s$ are optimized by gradient descent, while the teacher parameters $\theta_t$ are updated as an exponential moving average of $\theta_s$.

\paragraph{Profile-level distillation.}
We project the \texttt{[CLS]} representation of each view into a probability
distribution. Let $p_t^{(u)}$ and $p_s^{(v)}$ denote the teacher and student
outputs for two different views of the same profile. The profile-level
objective is
\begin{equation}
\mathcal{L}_{\mathrm{profile}}
=
\frac{1}{|\mathcal{P}_{\mathrm{profile}}|}
\sum_{(u,v)\in\mathcal{P}_{\mathrm{profile}}}
H\left(
\operatorname{sg}[p_t^{(u)}],
p_s^{(v)}
\right),
\end{equation}
where $\mathcal{P}_{\mathrm{profile}}$ contains valid cross-view pairs,
$\operatorname{sg}[\cdot]$ stops gradients through the teacher, and $H$ denotes
cross-entropy. This objective is averaged over both real and virtual profiles.
\begin{table}[t]
\centering
\caption{\textbf{Self-supervised pretraining configuration of PFE.} The model is trained using Distributed Data Parallel (DDP) on four NVIDIA RTX 4090 GPUs.
}
\label{tab:pfe_pretraining}
\small
\setlength{\tabcolsep}{5pt}
\begin{tabular}{lll}
\toprule
\textbf{Category} & \textbf{Hyperparameter} & \textbf{Value} \\
\midrule

\multirow{5}{*}{Expression views}
& Profile-wide views & 3 \\
& Profile-wide view composition & 1 clean + 2 posterized \\
& Subset-perturbed views & 6 \\
& Subset-perturbed Gaussian noise & $\sigma=0.05$ \\
& Gaussian perturbation ratio & 30\% of observed proteins \\
\midrule

\multirow{3}{*}{Protein masking}
& Profile-wide student mask ratios & 0.05 / 0.10 / 0.15 \\
& Teacher random masking & None \\
& Subset-perturbed mask ratios
& 0.07--0.20 \\
\midrule

\multirow{6}{*}{Virtual profiles}
& Virtual profiles / GPU / step & 64 \\
& Real profiles / GPU / step & 32 \\
& Augmentation types
& Posterization / Gaussian / mixup \\
& Augmentation probability
& $1/3$ each \\
& Gaussian noise & $\sigma=0.12$ \\
& Mixup coefficient
& $\lambda \sim \mathrm{Beta}(0.4,0.4)$ \\
\midrule

\multirow{7}{*}{Self-distillation}
& Profile loss weight & 1.0 \\
& Protein loss weight & 1.0 \\
& KoLeo weight & 0.1 \\
& Student temperature & 0.10 \\
& Teacher temperature
& $0.04 \rightarrow 0.07$ \\
& Temperature warm-up & 10 epochs \\
& Center momentum & 0.90 \\
\midrule

\multirow{8}{*}{Optimization}
& Optimizer & Adam \\
& Peak learning rate & $7\times10^{-5}$ \\
& Minimum learning rate & $1\times10^{-6}$ \\
& Learning-rate schedule & Warm-up + cosine decay \\
& Warm-up epochs & 10 \\
& Weight decay & $1\times10^{-5}$ \\
& Gradient clipping & 3.0 \\
& Epochs & 100 \\
\midrule

\multirow{2}{*}{EMA teacher}
& Initial momentum & 0.996 \\
& Momentum schedule & Cosine to 1.0 \\
\bottomrule
\end{tabular}
\end{table}
\paragraph{Protein-level distillation.}
For subset-perturbed student views, we additionally align each artificially
masked protein with its unmasked counterpart in a profile-wide teacher view.
Let $\mathcal{M}^{(v)}$ denote the masked protein positions in student view $v$.
The protein-level objective is
\begin{equation}
\mathcal{L}_{\mathrm{protein}}
=
\frac{1}{|\mathcal{P}_{\mathrm{protein}}|}
\sum_{(u,v)\in\mathcal{P}_{\mathrm{protein}}}
\frac{1}{|\mathcal{M}^{(v)}|}
\sum_{j\in\mathcal{M}^{(v)}}
H\left(
\operatorname{sg}[p_{t,j}^{(u)}],
p_{s,j}^{(v)}
\right),
\end{equation}
where $\mathcal{P}_{\mathrm{protein}}$ pairs profile-wide teacher views with
subset-perturbed student views. This objective is computed only for real
profiles.

\paragraph{Overall objective.}
The complete pretraining objective is
\begin{equation}
\mathcal{L}_{\mathrm{PFE}}
=
\mathcal{L}_{\mathrm{profile}}
+
\mathcal{L}_{\mathrm{protein}}
+
\lambda_{\mathrm{K}}\mathcal{L}_{\mathrm{KoLeo}},
\end{equation}
where $\mathcal{L}_{\mathrm{KoLeo}}$ discourages collapse among profile-level
representations. Virtual profiles contribute to
$\mathcal{L}_{\mathrm{profile}}$ but are excluded from
$\mathcal{L}_{\mathrm{protein}}$, preventing artificially edited protein values from serving as fine-grained token targets.

\subsection{PFE Pretraining Configuration}
\label{Configuration}

PFE uses a seven-layer Transformer with a hidden dimension of 768 and eight attention heads. Each input profile is represented by 38 protein tokens and one learnable \texttt{[CLS]} token, producing both profile-level and contextualized protein-level representations. Table~\ref{tab:pfe_architecture} summarizes the architectural configuration, including the continuous value encoder and self-distillation projection head.

We pretrain PFE for 100 epochs using distributed data parallelism on four NVIDIA RTX 4090 GPUs. Each GPU generates 64 virtual profiles per training step, and each real or virtual profile is used to construct three profile-wide views and six subset-perturbed views. Optimization uses Adam with a 10-epoch learning-rate warm-up followed by cosine decay. Detailed augmentation settings, self-distillation hyperparameters, and optimization schedules are provided in Table~\ref{tab:pfe_pretraining}. The compact panel of 38 cancer-related targets yields only 39 tokens per profile, keeping the memory overhead of self-attention modest and making pretraining feasible on four NVIDIA RTX 4090 GPUs under this configuration.

\begin{table}[t]
\centering
\caption{\textbf{Comparison of representative protein modeling approaches.} We summarize primary inputs, principal learning objectives, and representations or outputs.}
\label{tab:protein_model_comparison}
\small
\setlength{\tabcolsep}{5pt}
\renewcommand{\arraystretch}{1.15}
\begin{tabularx}{\linewidth}{@{}>{\raggedright\arraybackslash}p{0.18\linewidth}>{\raggedright\arraybackslash}X>{\raggedright\arraybackslash}X>{\raggedright\arraybackslash}X@{}}
\toprule
\textbf{Model} & \textbf{Primary Input} & \textbf{Learning Objective} & \textbf{Representation / Output} \\
\midrule
ESM-2 \citep{esm}
& Amino-acid sequences
& Masked amino-acid prediction
& Residue and protein embeddings \\

AlphaFold 2 \citep{jumper2021highly}
& Sequences, alignments, and structural templates
& Structure prediction
& Three-dimensional protein structures \\

AlphaFold 3 \citep{abramson2024accurate}
& Biomolecular sequences and ligand identities
& Diffusion-based structure prediction
& Three-dimensional biomolecular complexes \\

scPROTEIN \citep{li2024scprotein}
& Single-cell peptide and protein measurements
& Graph contrastive learning and uncertainty estimation
& Cell embeddings \\

scpFormer \citep{zhou2026scpformer}
& Single-cell protein abundance and sequence-derived embeddings
& Multi-objective masked expression reconstruction
& Cell and contextualized protein embeddings \\

VirTues \citep{wenckstern2026virtual}
& Multiplex protein images and sequence-derived marker embeddings
& Masked image reconstruction
& Marker, cell, niche, and tissue representations \\
\midrule
\textbf{PFE (Ours)}
& Bulk RPPA profiles
& Expression-space multi-view self-supervised learning
& Profile and contextualized protein embeddings \\
\bottomrule
\end{tabularx}
\end{table}

\subsection{Comparison of Protein Modeling Paradigms}
\label{ProteinModelingParadigms}

Protein representation learning encompasses amino-acid sequence modeling, molecular structure prediction, and the analysis of single-cell and spatial proteomic measurements. These approaches characterize individual proteins, cellular expression states, or spatial tissue organization, as summarized in Table~\ref{tab:protein_model_comparison}. Learning reusable representations directly from sample-level bulk protein expression remains an emerging direction.
Bulk proteomic pretraining faces practical constraints arising from limited cohort sizes and heterogeneous assay panels. To make effective use of available data, we integrate complementary RPPA resources into a shared protein panel and augment the resulting corpus through virtual profile generation. By perturbing selected expression values while preserving protein identities and natural missingness patterns, this strategy exposes PFE to additional local expression configurations anchored to observed samples.

Pretraining objectives also differ across modeling paradigms. ESM-2 and scpFormer employ masked prediction, whereas scPROTEIN uses graph contrastive learning. PFE adopts expression-space multi-view self-supervised learning to learn profile- and protein-level representations through self-distillation across perturbed views. While masked abundance reconstruction focuses on recovering withheld measurements, our objectives explicitly promote consistency under expression quantization, localized noise, and partial masking. This formulation encourages representations that remain stable under the specified perturbations, providing a proteomic reference space for subsequent Prot2Path distillation.

\begin{table}[t]
\centering
\caption{\textbf{Architectural configuration of ProSlide.}}
\label{tab:proslide_architecture}
\small
\setlength{\tabcolsep}{6pt}
\renewcommand{\arraystretch}{1.1}
\begin{tabular}{ll}
\toprule
\textbf{Parameter} & \textbf{Configuration} \\
\midrule
Patch encoder & CONCH v1.5 (frozen) \\
Patch feature dimension & 768 \\
Region grid & $4\times4$ patches \\
RSA depth & 3 \\
RSA attention heads & 8 \\
RSA MLP ratio & 2 \\
RSA aggregation & Learnable region \texttt{[CLS]} token \\
SSA depth & 7 \\
SSA attention heads & 8 \\
SSA MLP ratio & 4 \\
SSA aggregation & Learnable slide \texttt{[CLS]} token \\
Dropout & 0.1 \\
Attention dropout & 0.1 \\
Activation & GELU \\
Normalization & Pre-LN + final LayerNorm \\
Positional encoding & None \\
WSI representation & 768-dimensional slide \texttt{[CLS]} output \\
Protein regression heads & Independent linear layers: $768\rightarrow1$ \\
\bottomrule
\end{tabular}
\end{table}
\section{ProSlide Pretraining and Prot2Path Distillation}
\label{app:pathology_pretraining}

To facilitate reproducibility of Stages II and III, we detail the architecture of ProSlide, and the implementation of Prot2Path relational distillation.

\subsection{ProSlide Architecture}
\label{ProSlideArchitecture}

ProSlide follows a patch-to-region-to-slide hierarchy, with architectural settings summarized in Table~\ref{tab:proslide_architecture}. Using the patch-grid coordinates, we partition each WSI into non-overlapping regions spanning $4\times4$ patch positions. Within each region, the frozen CONCH v1.5 features are processed by a three-layer RSA encoder together with a learnable region \texttt{[CLS]} token. The corresponding output token serves as the region representation.

The region representations are processed by a seven-layer SSA encoder with a learnable slide \texttt{[CLS]} token, whose output forms the WSI representation. Both encoders use eight attention heads, GELU activations, pre-layer normalization, and a final LayerNorm. Patch-grid coordinates are used only to determine region membership, while neither RSA nor SSA introduces positional embeddings.

For protein-expression supervision, each target is assigned an independent linear regression head mapping the WSI representation to a scalar expression value. The feature projections, RSA and SSA encoders, learnable \texttt{[CLS]} tokens, and regression heads are jointly trained, while the CONCH v1.5 patch encoder remains frozen.

\subsection{Prot2Path Implementation Details}
\label{Prot2PathImplementation}

To establish a shared molecular reference space for distillation, we first encode the candidate protein profiles using the frozen PFE. For each profile, we extract the 768-dimensional \texttt{[CLS]} representation after the final LayerNorm, without applying the pretraining projection head. The candidate reference pool contains 12,695 real protein profiles: 1,338 paired training profiles and 11,357 standalone profiles. Table~\ref{tab:prot2path_implementation} summarizes the implementation settings.
A linear alignment head with bias maps the ProSlide representation into the 768-dimensional PFE space. Teacher representations, projected WSI representations, and bank entries are $L_2$-normalized before computing cosine similarities. Both teacher and student temperatures are fixed at 0.05. We minimize the teacher-to-student KL divergence, summing over reference entries and averaging over paired samples in each mini-batch. The loss has unit weight, with no temperature-squared scaling or auxiliary training losses.

\begin{table}[t]
\centering
\caption{\textbf{Implementation configuration of Prot2Path.}}
\label{tab:prot2path_implementation}
\small
\setlength{\tabcolsep}{6pt}
\renewcommand{\arraystretch}{1.1}
\begin{tabular}{ll}
\toprule
\textbf{Parameter} & \textbf{Configuration} \\
\midrule
Proteomic teacher & Frozen PFE \\
Teacher representation & Final \texttt{[CLS]} output after LayerNorm \\
Representation dimension & 768 \\
Candidate pool size & 12,695 real profiles \\
Sub-bank size
    & $K_{\mathrm{sub}} = 1{,}594$ \\
Paired references
    & All 1,338 paired training profiles \\
Standalone references
    & 256 randomly sampled standalone profiles \\
Representation normalization & $L_2$ normalization \\
Similarity measure & Cosine similarity \\
Teacher temperature & $\tau_{\mathrm{t}}=0.05$ \\
Student temperature & $\tau_{\mathrm{s}}=0.05$ \\
Distillation objective & Teacher-to-student KL divergence \\
\bottomrule
\end{tabular}
\end{table}

\section{Downstream Evaluation and Additional Analyses}
\label{app:downstream}

To complement the downstream evaluation in the main text, we provide dataset and task descriptions, baseline configurations, and detailed results. We also present additional analyses of pretraining data efficiency, GPU memory usage during inference, and model interpretability.

\subsection{Datasets and Task Definitions}
\label{app:Datasets}

To assess the downstream utility of proteomics-guided pathology representations, we consider classification endpoints that characterize both molecular states and tissue morphology. The benchmark includes mutation prediction, receptor-status prediction, and histological subtyping, with cross-cohort settings incorporated to assess transferability across data sources. Below, we detail the 12 WSI-level tasks, grouped into renal (Tasks 1--4), breast (Tasks 5--8), and lung (Tasks 9--12) cancer benchmarks.
\begin{itemize}

\item \textbf{Task 1 (MUT-SETD2).}
This task predicts \textit{SETD2} mutation status in the MUT-HETRCC cohort. It evaluates morphology associated with this chromatin-regulation gene alteration.

\item \textbf{Task 2 (MUT-BAP1).}
This task predicts \textit{BAP1} mutation status in the MUT-HETRCC cohort. It provides an additional renal-cancer mutation benchmark outside the CPTAC cohort.

\item \textbf{Task 3 (CPTAC-CCRCC-BAP1).}
This CPTAC clear-cell renal-cell carcinoma task predicts \textit{BAP1} mutation status. It tests the relationship between WSI morphology and a key renal-cancer molecular alteration.

\item \textbf{Task 4 (Cross-MUT-BAP1).}
This task predicts \textit{BAP1} mutation status under cross-cohort transfer. Patients from MUT-HETRCC are split into training and validation sets, while CPTAC-CCRCC serves as the fixed external test cohort. This setting evaluates whether a classifier trained on MUT-HETRCC transfers to CPTAC-CCRCC.

\item \textbf{Task 5 (CPTAC-BRCA-PR).}
This task uses the CPTAC breast cancer cohort to predict progesterone-receptor (PR) status from whole-slide images. It evaluates whether histomorphological patterns are informative for this clinically relevant receptor phenotype.

\item \textbf{Task 6 (CPTAC-BRCA-HER2).}
This CPTAC breast cancer task predicts \textit{HER2} status. It assesses whether slide-level representations capture histomorphological patterns associated with HER2-positive breast cancer.

\item \textbf{Task 7 (CPTAC-BRCA-TP53).}
This task predicts \textit{TP53} mutation status in the CPTAC breast cancer cohort. It serves as a molecular-phenotype benchmark with potentially heterogeneous mutation-associated morphology.

\item \textbf{Task 8 (BCNB-ER).}
This task predicts estrogen-receptor (\textit{ER}) status in the BCNB breast cancer cohort. It complements the CPTAC breast cancer tasks with an independent data source.

\item \textbf{Task 9 (CPTAC-LUAD-EGFR).}
This CPTAC lung adenocarcinoma task predicts \textit{EGFR} mutation status from whole-slide images. It evaluates sensitivity to a clinically actionable genomic alteration in lung cancer.

\item \textbf{Task 10 (Combine-LUNG fine subtype).}
This task performs three-class classification of the predominant histological patterns of lung adenocarcinoma: acinar, lepidic, and solid. Samples from the Local-LUNG and DHMC-LUNG cohorts are pooled to evaluate fine-grained histological subtype discrimination using multi-source data.

\item \textbf{Task 11 (Cross-LUNG subtyping).}
This task distinguishes lung adenocarcinoma (LUAD) from lung squamous cell carcinoma (LSCC) under cross-cohort transfer. Local-LUNG provides the training and validation patients, while DHMC-LUNG, CPTAC-LUAD, and CPTAC-LSCC jointly form the fixed external test set.

\item \textbf{Task 12 (Cross-LUNG fine subtype).}
This task performs three-class classification of the predominant histological patterns of lung adenocarcinoma: acinar, lepidic, and solid. The Local-LUNG cohort is used for training and validation, while DHMC-LUNG serves as the fixed test cohort. This setting evaluates cross-cohort generalization in fine-grained histological subtype discrimination.

\end{itemize}

\begin{table}[t]
\centering
\caption{Task-specific cohort sizes for downstream evaluation.}
\label{tab:datasetsintro}
\small
\setlength{\tabcolsep}{6pt}
\renewcommand{\arraystretch}{1.1}
\begin{tabular}{lllr}
\toprule
Task & Cohort(s) & Patients ($N_p$) & WSIs ($N_s$) \\
\midrule
Task 1 & MUT-HETRCC
        & 1292 & 1292 \\
Task 2 & MUT-HETRCC
        & 1292 & 1292 \\
Task 3 & CPTAC-CCRCC
        & 103 & 245 \\
Task 4 & MUT-HETRCC + CPTAC-CCRCC
       & 1395 & 1537 \\
\midrule
Task 5 & CPTAC-BRCA
        & 97 & 106 \\
Task 6 & CPTAC-BRCA
        & 82 & 89 \\
Task 7 & CPTAC-BRCA
        & 103 & 112 \\
Task 8 & BCNB
        & 1028 & 1028 \\
\midrule
Task 9 & CPTAC-LUAD
        & 108 & 324 \\
Task 10 & Local-LUNG + DHMC-LUNG
         & 395 & 410 \\
Task 11 & Local-LUNG + CPTAC-(LUAD \& LSCC) + DHMC-LUNG
         & 1003 & 2819 \\
Task 12 & Local-LUNG + DHMC-LUNG
         & 395 & 410 \\
\bottomrule
\end{tabular}
\end{table}

\begin{table*}[t]
\centering
\caption{\textbf{Comparison of representative slide-level pathology foundation models.}
We compare the patch encoder, whole-slide aggregation strategy, pretraining scale, and supervision of the pathology foundation models considered in this work.}
\label{tab:wsi_fm_comparison}
\small
\setlength{\tabcolsep}{4pt}
\renewcommand{\arraystretch}{1.12}

\begin{tabular}{
lllll}
\toprule
\textbf{Model} &
\textbf{Patch Encoder} &
\textbf{Slide Aggregation} &
\textbf{Pretraining Scale} &
\textbf{Pretrain Sup.} \\

\midrule

CARE
& CONCH v1.5
& CARE
& 34,277 WSIs
& iBoT \& Clip
\\

TITAN
& CONCH v1.5
& ViT
& 335,645 WSIs
& iBoT \& Coca \\

PRISM
& Virchow
& Perceiver
& 587,196 WSIs
& Coca \\

GigaPath
& GigaPath
& LongNet
& 171,189 WSIs
& MAE \\

FEATHER
& CONCH v1.5
& ABMIL
& 24,000 WSIs
& Supervised classification\\
\midrule
\multirow{2}{*}{ProSlide}
& \multirow{2}{*}{CONCH v1.5}
& \multirow{2}{*}{Region ViT}
& {12,695 protein profiles} 
& {Protein supervision \& }  \\
&&& { \& 2,229 WSIs}&{Proteomic knowledge distill}\\
\bottomrule
\end{tabular}
\end{table*}

The benchmark comprises nine molecular prediction tasks (Tasks 1--9), covering gene mutation and receptor-status prediction, and three histological subtyping tasks (Tasks 10--12). Task 10 uses pooled data from Local-LUNG and DHMC-LUNG, while Tasks 4, 11, and 12 assess cross-cohort transfer across different data sources.
We use repeated patient-level holdout splits (Table \ref{tab:datasetsintro}). All WSIs from the same patient were assigned to the same partition within each split. Tasks~1, 2, 8, and~10 used five repeated patient-level splits, with target allocations of 60\% for training, 20\% for validation, and 20\% for testing. Tasks~3, 5, 6, 7, and~9 used 50 repeated patient-level splits, allocating 80\% of patients to training and 20\% to testing, without a separate validation set. For Tasks~4, 11, and~12, the test patients were fixed across all five repetitions, while the development pool was repartitioned into training and validation sets at a 3:1 ratio. This ratio applies only to the development pool and excludes the fixed test set. The split proportions refer to patients rather than WSIs. Because patients can contribute different numbers of WSIs, slide counts need not follow the same proportions. Integer partition sizes can also differ slightly from the nominal patient-level proportions.

\subsection{Evaluation Protocols and Baselines}
\label{app:Evaluation}

To assess the transferability of slide representations learned through different pretraining strategies, we evaluate all models using the frozen-feature linear-probing protocol described in the main text. Table~\ref{tab:wsi_fm_comparison} summarizes their patch encoders, slide aggregation architectures, and slide-level pretraining settings.
The multimodal baselines incorporate molecular or textual supervision. CARE \citep{zhang2026care} combines adaptive region modeling with iBOT pretraining and subsequent contrastive alignment with transcriptomic and protein profiles. TITAN \citep{titan} uses a slide-level Vision Transformer pretrained through iBOT and CoCa-based alignment with synthetic captions and pathology reports. PRISM \citep{prism} aggregates Virchow patch features using a Perceiver and jointly optimizes contrastive alignment and report generation.
The remaining baselines represent visual self-supervision and supervised pretraining. GigaPath \citep{gigapath} uses a LongNet-based slide encoder pretrained through masked autoencoding. FEATHER \citep{feather} learns transferable slide representations through attention-based multiple instance learning and pan-cancer classification. We consider its CONCH v1.5-based FEATHER-24K variant, pretrained on approximately 24,000 slides across 108 morphological categories.

\begin{table*}[t]
\centering
\caption{Logistic-regression results on the selected 12-task set. Values are mean$\pm$standard deviation. The best and second-best means within each task-metric row are shown in bold and underlined, respectively.}
\label{tab:pancancer-12task-lr}
\scriptsize
\setlength{\tabcolsep}{2.8pt}
\resizebox{\textwidth}{!}{%
\renewcommand{\arraystretch}{1.2}
\begin{tabular}{@{}ll*{7}{c}@{}}
\toprule
Task & Metric
& \multicolumn{1}{c}{Mean-pooling}
& \multicolumn{1}{c}{FEATHER}
& \multicolumn{1}{c}{GigaPath}
& \multicolumn{1}{c}{PRISM}
& \multicolumn{1}{c}{CARE}
& \multicolumn{1}{c}{TITAN}
& \multicolumn{1}{c}{ProSlide} \\
\midrule

\multirow{2}{*}{Task 1}
& ACC
& 54.52$\pm$0.04
& 54.88$\pm$0.07
& 55.98$\pm$0.04
& 55.93$\pm$0.11
& \underline{56.72$\pm$0.02}
& 56.29$\pm$0.07
& \textbf{61.65$\pm$0.20} \\
& AUC
& 70.32$\pm$0.13
& 73.41$\pm$0.10
& 72.71$\pm$0.05
& 72.29$\pm$0.10
& \underline{73.54$\pm$0.08}
& 72.56$\pm$0.06
& \textbf{76.09$\pm$0.07} \\

\multirow{2}{*}{Task 2}
& ACC
& 60.95$\pm$0.04
& 54.74$\pm$0.03
& \underline{61.44$\pm$0.13}
& 57.55$\pm$0.15
& \underline{61.44$\pm$0.07}
& 59.83$\pm$0.14
& \textbf{62.87$\pm$0.16} \\
& AUC
& 84.84$\pm$0.06
& 84.28$\pm$0.05
& 86.57$\pm$0.01
& 86.49$\pm$0.01
& \textbf{88.89$\pm$0.02}
& 85.98$\pm$0.01
& \underline{86.89$\pm$0.03} \\

\multirow{2}{*}{Task 3}
& ACC
& 57.50$\pm$0.83
& 58.14$\pm$1.61
& 55.18$\pm$0.65
& 53.98$\pm$1.13
& 58.70$\pm$1.03
& \textbf{61.20$\pm$1.43}
& \underline{60.39$\pm$1.14} \\
& AUC
& 62.61$\pm$2.92
& 58.23$\pm$3.68
& 60.08$\pm$2.79
& 59.52$\pm$2.78
& 65.83$\pm$2.61
& \textbf{73.16$\pm$2.08}
& \underline{67.69$\pm$2.31} \\

\multirow{2}{*}{Task 4}
& ACC
& \underline{58.30$\pm$0.08}
& 52.03$\pm$0.03
& 51.57$\pm$0.16
& 57.64$\pm$0.05
& 56.87$\pm$0.10
& 57.35$\pm$0.16
& \textbf{58.48$\pm$0.32} \\
& AUC
& 70.87$\pm$0.08
& 67.90$\pm$0.19
& \textbf{81.96$\pm$0.02}
& 72.58$\pm$0.01
& 72.68$\pm$0.05
& 71.24$\pm$0.01
& \underline{76.00$\pm$0.03} \\

\multirow{2}{*}{Task 5}
& ACC
& 63.06$\pm$0.97
& \underline{65.95$\pm$1.34}
& \textbf{68.57$\pm$1.03}
& 63.10$\pm$1.20
& 57.51$\pm$1.07
& 59.82$\pm$0.89
& 65.63$\pm$0.73 \\
& AUC
& 69.57$\pm$1.27
& 73.05$\pm$1.27
& \textbf{75.65$\pm$1.72}
& 72.17$\pm$1.15
& 62.10$\pm$1.56
& 68.68$\pm$1.27
& \underline{73.82$\pm$1.15} \\

\multirow{2}{*}{Task 6}
& ACC
& 63.11$\pm$1.51
& 59.65$\pm$1.25
& 62.04$\pm$0.86
& 54.77$\pm$1.21
& \underline{64.52$\pm$0.76}
& 62.79$\pm$1.00
& \textbf{64.54$\pm$1.29} \\
& AUC
& \underline{69.14$\pm$1.51}
& 63.85$\pm$1.21
& \textbf{72.16$\pm$1.39}
& 59.28$\pm$1.51
& 66.60$\pm$1.13
& 68.14$\pm$1.09
& 68.38$\pm$1.43 \\

\multirow{2}{*}{Task 7}
& ACC
& 74.63$\pm$0.89
& 67.54$\pm$0.85
& 67.05$\pm$0.85
& 73.18$\pm$0.96
& 74.08$\pm$0.74
& \textbf{78.22$\pm$0.67}
& \underline{75.39$\pm$0.90} \\
& AUC
& 81.23$\pm$1.02
& 75.60$\pm$0.65
& 74.68$\pm$0.98
& 79.84$\pm$0.85
& 83.19$\pm$0.56
& \textbf{87.70$\pm$0.44}
& \underline{84.43$\pm$0.79} \\

\multirow{2}{*}{Task 8}
& ACC
& 70.16$\pm$0.12
& 74.02$\pm$0.23
& 71.11$\pm$0.61
& \underline{74.50$\pm$0.16}
& 74.39$\pm$0.12
& \textbf{74.52$\pm$0.09}
& \textbf{74.52$\pm$0.21} \\
& AUC
& 82.85$\pm$0.23
& 86.69$\pm$0.16
& 87.51$\pm$0.21
& \underline{88.24$\pm$0.10}
& \textbf{88.40$\pm$0.08}
& 88.22$\pm$0.08
& 87.50$\pm$0.10 \\

\multirow{2}{*}{Task 9}
& ACC
& 69.97$\pm$0.54
& 69.46$\pm$0.49
& \textbf{74.17$\pm$0.73}
& 69.34$\pm$0.53
& 72.25$\pm$0.50
& 68.94$\pm$0.64
& \underline{72.93$\pm$0.65} \\
& AUC
& 76.81$\pm$0.69
& 76.47$\pm$0.59
& \textbf{81.14$\pm$0.62}
& 77.32$\pm$0.56
& \underline{79.75$\pm$0.48}
& 76.58$\pm$0.60
& 79.55$\pm$0.70 \\

\multirow{2}{*}{Task 10}
& ACC
& 52.40$\pm$0.16
& 59.90$\pm$0.07
& 49.41$\pm$0.44
& 58.17$\pm$0.01
& 57.53$\pm$0.28
& \textbf{60.90$\pm$0.21}
& \underline{60.53$\pm$0.15} \\
& AUC
& 72.65$\pm$0.06
& \textbf{77.39$\pm$0.06}
& 69.59$\pm$0.36
& 75.66$\pm$0.01
& 76.83$\pm$0.13
& 77.24$\pm$0.13
& \underline{77.28$\pm$0.14} \\

\multirow{2}{*}{Task 11}
& ACC
& 52.50$\pm$0.00
& 55.28$\pm$0.06
& \underline{57.91$\pm$0.10}
& \textbf{61.45$\pm$0.01}
& 52.46$\pm$0.00
& 55.12$\pm$0.10
& 53.51$\pm$0.01 \\
& AUC
& 78.06$\pm$0.01
& 76.76$\pm$0.12
& \textbf{80.83$\pm$0.01}
& 78.47$\pm$0.01
& 77.00$\pm$0.01
& 76.60$\pm$0.02
& \underline{79.10$\pm$0.11} \\

\multirow{2}{*}{Task 12}
& ACC
& 63.56$\pm$0.08
& 56.22$\pm$0.26
& 54.22$\pm$0.14
& 58.34$\pm$0.09
& \underline{65.53$\pm$0.03}
& 63.75$\pm$0.07
& \textbf{65.69$\pm$0.01} \\
& AUC
& 73.70$\pm$0.03
& 68.48$\pm$0.28
& 62.16$\pm$0.07
& 67.98$\pm$0.10
& \underline{74.37$\pm$0.00}
& 74.18$\pm$0.02
& \textbf{75.65$\pm$0.01} \\
\bottomrule
\end{tabular}%
}
\end{table*}

\subsection{Detailed Downstream Results}
\label{app:Results}

To examine performance beyond the aggregate comparisons, we provide task-level classification results.

\paragraph{Task-level performance.}
Table~\ref{tab:pancancer-12task-lr} reports the complete linear-probing results across the 12 downstream tasks. ProSlide has achieved competitive performance across molecular prediction and histological subtyping, with the highest ACC and AUC on both SETD2 mutation prediction and cross-cohort fine-grained lung subtyping. Its relative performance has nevertheless varied across endpoints, with other encoders leading on several CPTAC molecular tasks. Cross-cohort binary lung subtyping has also revealed a discrepancy between its competitive AUC and weaker ACC, highlighting the importance of considering both metrics when assessing transferability.


\subsection{Data Efficiency and Inference Memory Usage}
\label{app:Efficiency}

\begin{figure}[H]
	\begin{center}
		\includegraphics[width=\linewidth]{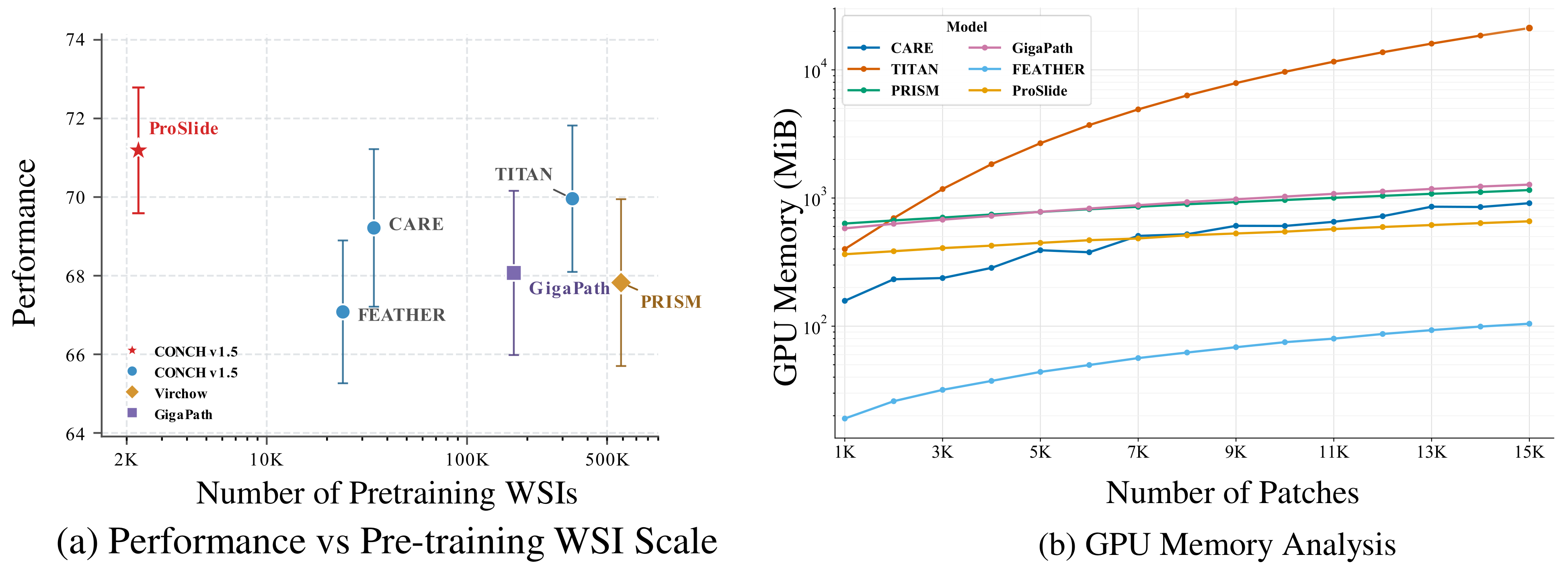}
	\end{center}
	\caption{\textbf{Data and memory efficiency.} ProSlide achieves strong performance with limited slide-level pretraining data (a) and low GPU memory usage during inference (b).}
	\label{model_analysis}
\end{figure}

\paragraph{Pretraining data efficiency.}
To examine downstream performance in relation to slide-level pretraining
scale, we compare the evaluated models in Figure~\ref{model_analysis}a.
ProSlide achieves the highest aggregate performance among these models
with 2,229 WSIs for slide-level pretraining. This comparison highlights
the potential of protein-derived supervision to support slide
representation learning with limited paired data. The reported WSI
counts refer specifically to slide-level pretraining; ProSlide also
draws on pretrained CONCH v1.5 features and PFE-derived proteomic
knowledge.

\paragraph{Inference memory usage.}
To examine inference memory requirements across slide sizes, we measure
GPU memory usage as the number of input patches increases
(Figure~\ref{model_analysis}b). ProSlide exhibits modest memory growth
over the evaluated range. Its memory usage remains below that of TITAN,
PRISM, and GigaPath, and falls below CARE at larger patch counts.
FEATHER has the lowest memory usage among the evaluated models.
ProSlide's memory behavior is consistent with its hierarchical
aggregation, which summarizes patch features into region
representations before slide-level attention.


\begin{figure}[H]
\centering
\includegraphics[width=0.9\linewidth]{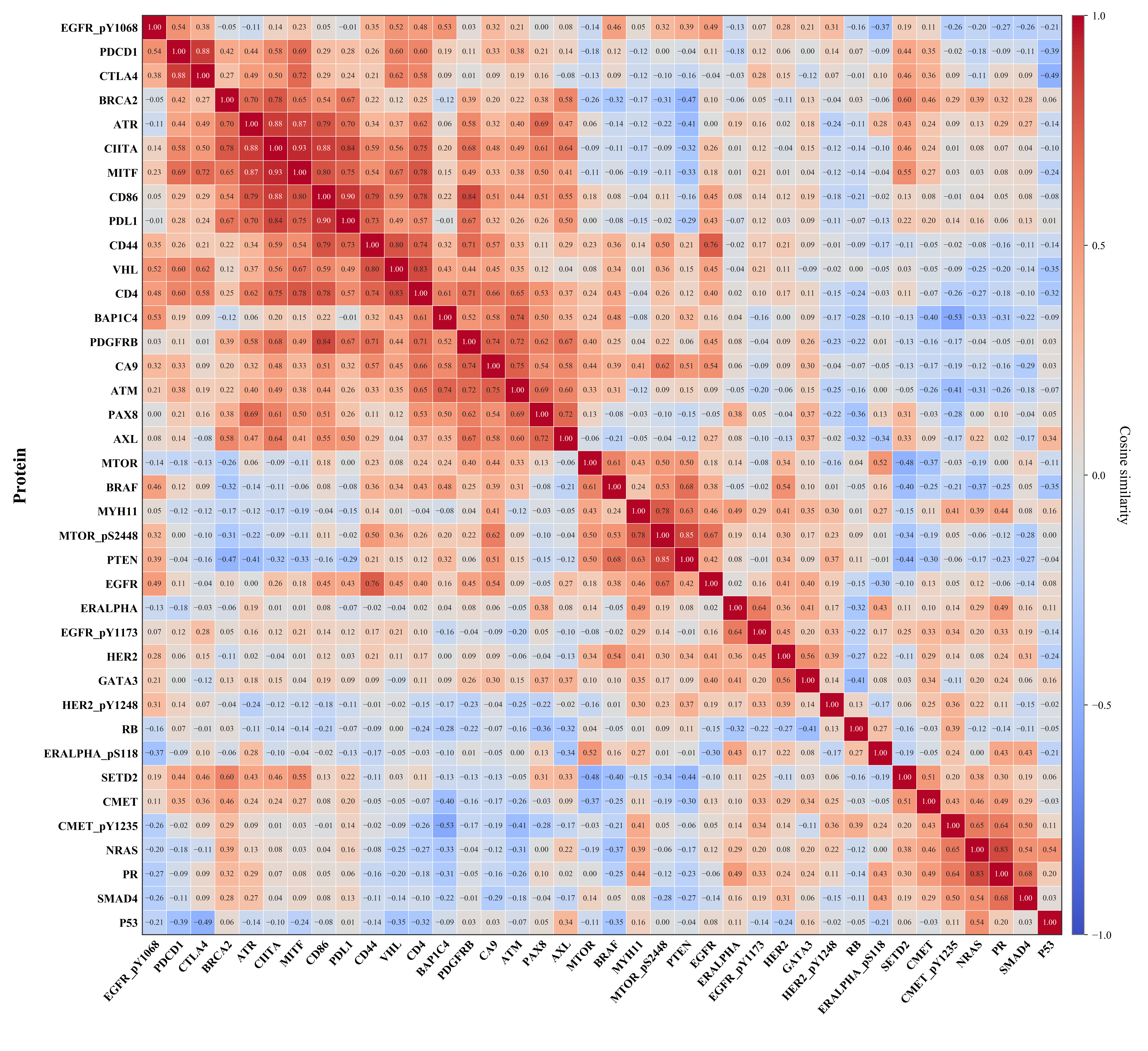}
\caption{\textbf{PFE protein identity embeddings.} Pairwise cosine similarities among 38 learned protein identity embeddings. Both axes follow the same clustering order, with warmer colors indicating higher similarity.}
\label{identifyemb}
\end{figure}

\subsection{Analysis of PFE Protein Identity Embeddings}
\label{app:Interpretability}

To assess whether the learned protein identity embeddings reflect known biological relationships, we  compute pairwise cosine similarities among the 38 target embeddings and visualize them in a clustered heatmap (Figure~\ref{identifyemb}). These learned input embeddings are shared across samples and provide the identity component of each protein token. Both axes follow the same clustering order to facilitate inspection of the resulting similarity structure.

The heatmap has shown high similarity between PDCD1 and CTLA4, both of which encode inhibitory immune-checkpoint receptors involved in T-cell regulation \citep{pardoll2012blockade}. BRCA2 and ATR also exhibit relatively high similarity. Their functional connection involves DNA damage signaling and repair, with ATM/ATR-mediated PALB2 phosphorylation promoting RAD51 function within the PALB2--BRCA2 repair pathway \citep{ahlskog2016atm}. These examples  illustrate a qualitative correspondence between selected embedding similarities and known biological relationships.

\end{document}